%% file: main.tex
\documentclass[10pt,twocolumn,letterpaper]{article}

\usepackage[pagenumbers]{cvpr} 

\input{preamble}

\definecolor{cvprblue}{rgb}{0.21,0.49,0.74}
\usepackage[pagebackref,breaklinks,colorlinks,citecolor=cvprblue]{hyperref}

\def\paperID{2} 
\def\confName{CVPRW}
\def\confYear{2024}

\title{Divide and Conquer: High-Resolution Industrial Anomaly Detection via Memory Efficient Tiled Ensemble}

\author{
Blaž Rolih$^1$
\and
Dick Ameln$^2$
\and
Ashwin Vaidya$^2$
\and
Samet Akcay$^2$
\and
$ ^{1}$University of Ljubljana, Faculty of Computer and Information Science \\
$ ^{2}$Intel \\
{\tt\small br9136@student.uni-lj.si},
{\tt\small \{dick.ameln, ashwin.vaidya, samet.akcay\}@intel.com}
}

\begin{document}
\maketitle
\input{sec/0_abstract}    
\input{sec/1_intro}
\input{sec/2_related_work}
\input{sec/3_method}
\input{sec/4_experiments}
\input{sec/5_results}
\input{sec/6_conclusion}
{
    \small
    \bibliographystyle{ieeenat_fullname}
    \bibliography{main}
}

\input{sec/X_suppl}

\end{document}

%% file: preamble.tex
\usepackage{makecell}

\usepackage[dvipsnames]{xcolor}

\usepackage[accsupp]{axessibility}
\usepackage{placeins}

\usepackage{tikz}
\definecolor{gold}{HTML}{FBF2D2}
\definecolor{silver}{HTML}{DDDDDD}
\definecolor{bronze}{HTML}{EED2B8}

\definecolor{goldD}{HTML}{D9AE13}
\definecolor{silverD}{HTML}{909090}
\definecolor{bronzeD}{HTML}{9A5F26}

\newcommand{\medal}[3]{\tikz[baseline=(char.base)]{\node[rounded corners=1pt,fill=#1,draw=#2,inner sep=0pt](char){#3};}}

\newcommand{\bm}[2]{
    \ifcase#1\or
      {\medal{gold}{goldD}{\textbf{#2}}}
    \or 
      {\medal{silver}{silverD}{#2}}
    \or 
      {\medal{bronze}{bronzeD}{#2}}
    \else 
      #2
    \fi\ignorespaces
}

%% file: sec/0_abstract.tex
\begin{abstract}

Industrial anomaly detection is an important task within computer vision with a wide range of practical use cases. The small size of anomalous regions in many real-world datasets necessitates processing the images at a high resolution. This frequently poses significant challenges concerning memory consumption during the model training and inference stages, leaving some existing methods impractical for widespread adoption.
To overcome this challenge, we present the tiled ensemble approach, which reduces memory consumption by dividing the input images into a grid of tiles and training a dedicated model for each tile location. The tiled ensemble is compatible with any existing anomaly detection model without the need for any modification of the underlying architecture.
By introducing overlapping tiles, we utilize the benefits of traditional stacking ensembles, leading to further improvements in anomaly detection capabilities beyond high resolution alone. We perform a comprehensive analysis using diverse underlying architectures, including Padim, PatchCore, FastFlow, and Reverse Distillation, on two standard anomaly detection datasets: MVTec and VisA. Our method demonstrates a notable improvement across setups while remaining within GPU memory constraints, consuming only as much GPU memory as a single model needs to process a single tile.
\footnote{Available as part of Anomalib: \\ \url{https://github.com/openvinotoolkit/anomalib}.}
\footnote{Research conducted during GSoC 2023 at OpenVINO.}

\end{abstract}

%% file: sec/1_intro.tex
\section{Introduction}
\label{sec:intro}

The detection and localization of anomalies in images is a crucial task with a wide range of industrial applications. The ability to identify hard-to-detect defects of various sizes within images enables automation of many processes, maintenance of safety, and prevention of financial loss.

\begin{figure}[t]
  \centering
   \includegraphics[width=1\linewidth]{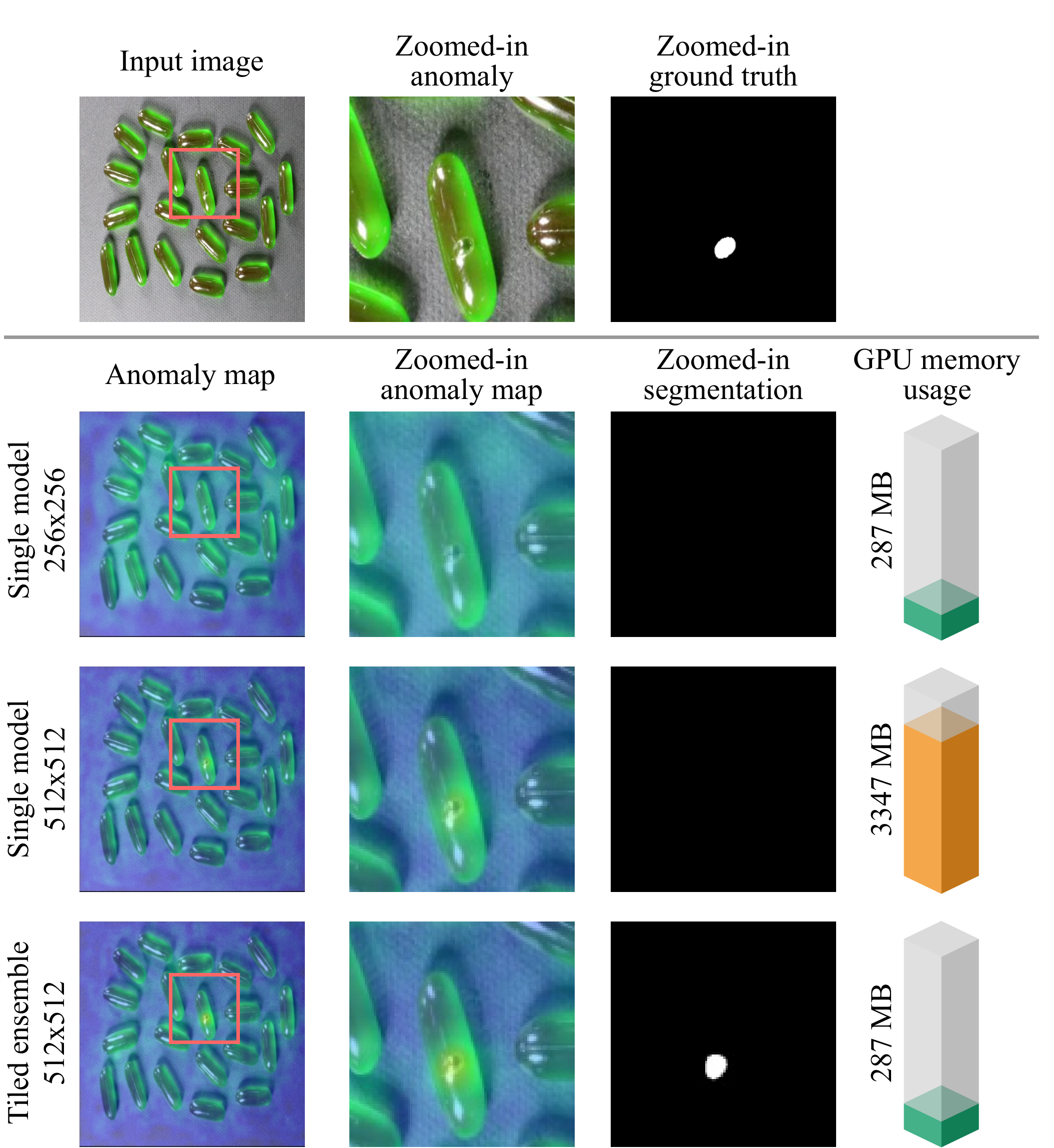}
   \caption{Example of anomaly localization on VisA capsules category. A tiled ensemble successfully manages to detect an anomaly, while a single model with a smaller resolution or equivalent resolution fails to do so. The tiled ensemble achieves this while remaining within the GPU memory constraints.}
   \label{fig:main_fig}
\end{figure}

In recent years, the field has witnessed substantial performance improvements, primarily driven by advancements in deep learning techniques. In addition, the high real-world potential of anomaly detection techniques has prompted a shift of focus towards efficiency, as latency, throughput, and memory consumption are important metrics to optimize when deploying models on resource-limited devices~\cite{liu_simplenet, gudovskiy_cflow, yu_fastflow, batzner_efficientad, lee_memory, park_fastano}.

A challenge of real-world datasets is that the size of the anomalous regions within the images may be very small relative to the full image size. The common practice of downscaling the input images to a predetermined input size may in such cases lead to a loss of pixel-level information, which in turn causes the model to miss small anomalies and incorrectly mark images as defect-free. Processing the images at their original resolution or reducing the amount of downscaling may constitute a natural tactic to prevent this type of false negative, but will at the same time inflate the memory consumption of the model. As a result, memory constraints may prevent processing the images in a resolution suitable for detecting the smallest anomalies in the dataset, especially in low-resource settings.

Tiling mechanisms, which subdivide the input images into a rectangular grid of tiles as a pre-processing step, have been used to process images at a high resolution while keeping memory use low~\cite{nejat_pathoAD}. By passing individual tiles to the model as input instead of full images, tiling reduces the model's input dimensions, while maintaining the effective input resolution of the images content-wise. This approach may not always be ideal, particularly for methods sensitive to object alignment~\cite{li_cutpaste}, as using a single model for all patches may compromise spatial information preservation.

Contrary to traditional tiling, our tiled ensemble trains a separate model on each of the individual tile locations. The full training procedure yields an ensemble of independently trained models, each specialized on a single specific tile location. By assigning a separate model to each tile location, we achieve a direct spatial mapping of feature space to pixel space, making the method suitable for spatially-aware models. An additional advantage of using separate models is that it allows us to leverage the benefits of stacking ensemble methods by introducing overlapping tiles, which further improves anomaly detection performance. By merging the predictions of the individual models as a post-processing step, we obtain an end-to-end pipeline, offering direct application to practical settings while ensuring that the peak GPU memory usage remains in the range of that required by a model processing a single tile. Since our approach only changes the pre- and post-processing stages, it is not limited to a specific model architecture and can be applied as an extension to any anomaly detection pipeline. \Cref{fig:main_fig} illustrates how a tiled ensemble can detect small anomalies by utilizing increased resolution without consuming excessive GPU memory. 

To showcase the applicability of our tiled ensemble, we benchmark the approach against non-tiling baselines, as well as traditional tiling methods, using a diverse set of architectures such as probability density modelling (Padim~\cite{defard_padim}), memory bank based (Patchcore~\cite{roth_patchcore}), student-teacher (Reverse Distillation~\cite{deng_rev_dist}), and normalizing flows (Fastflow~\cite{yu_fastflow}). For evaluation, we use two well-known anomaly detection datasets: MVTec AD~\cite{bergmann_mvtec} and VisA~\cite{zou_visa}, with an emphasis on detecting smaller anomalies, particularly evident in the VisA dataset.

\noindent Overall, this paper provides the following contributions:
\begin{itemize}
    \item We propose a practical approach for the detection and localization of anomalies in high-resolution images while adhering to GPU memory constraints. This enables the detection of small anomalies in real-world applications, increasing reliability and performance.

    \item Our approach offers a model-agnostic framework that can enhance both existing and upcoming anomaly detection architectures. By adopting our methodology, these architectures can better handle higher-resolution images, additionally benefiting in constrained settings without the need for modification of the underlying architecture.

    \item Having a dedicated model for each tile location enables the model to highly specialize in specific part of an image. Additionally, the integration of overlapping tiles in our approach has the advantages of conventional stacking ensembles. This results in enhanced performance that surpasses what can be achieved solely through increased resolution.
\end{itemize}    

%% file: sec/2_related_work.tex
\section{Related work}
\label{sec:related}

In recent times, there have been notable advancements in the field of visual anomaly detection, with numerous techniques being introduced based on various approaches such as reconstructive methods~\cite{zavrtanik_riad, bergmann_mvtecloco, bergmann_ssim-ae}, student-teacher networks~\cite{wang_stfpm, rudolph_ast, batzner_efficientad, deng_rev_dist, bergmann_uninformed}, discriminative methods~\cite{yang_memseg, zhang_destseg, zavrtanik_draem, zavrtanik_dsr, li_cutpaste, fuvcka_transfusion}, normalizing flows~\cite{yu_fastflow, rudolph_csflow, gudovskiy_cflow}, and embedding-based methods~\cite{defard_padim, roth_patchcore, liu_simplenet}.

Apart from the design of novel model architectures, an active area of focus has been enhancing and extending existing approaches. Recent research has indicated that the performance of anomaly detection models can benefit from careful design choices around data augmentation\cite{wang_unicon_pre}, pre-training characteristics\cite{he_dense_pre}, and feature space selection~\cite{heckler_backbones}. Other studies have focused on modifying or extending the architecture of existing models. \citeauthor{ristea_sspcab}~\cite{ristea_sspcab} introduced the SSPCAB block, which can be injected into various state-of-the-art methods to enhance their performance. Similarly, \citeauthor{simoes_attentionAD}~\cite{simoes_attentionAD} demonstrated that significant improvements can be achieved by extending existing architectures with attention blocks. Finally, \citeauthor{heckler_AD_featSel}~\cite{heckler_AD_featSel} developed a feature selection method that optimally selects a layer from the pre-trained feature extractor depending on the characteristics of the task. The tiled ensemble method follows a similar strategy, altering the pre- and post-processing stages of existing anomaly detection pipelines with the aim of improving the performance on datasets with small anomalies.

A common class of anomaly detection models is patch-based models. A patch refers to a spatial location in the intermediate feature embedding map of the model backbone and usually translates directly to a pixel area in the input images. Patch-based models aim to find the natural distribution of each patch location from the feature embeddings of normal images during training and estimate the distance of the feature embeddings to this distribution during inference. This process yields a set of patch-level anomaly scores which form the basis of the anomaly localization predictions. To find the distribution of a given patch location, a model may rely only on the embeddings of that same patch location~\cite{defard_padim, yi_patch_svdd, sohn_DisAugCLR, chen_OCC_IGD}, or alternatively also consider the interrelation among patches~\cite{tsai_multi_patch, roth_patchcore, park_fastano}. Further, the authors of CutPaste~\cite{li_cutpaste} discovered that employing separate models for each patch location yields superior results. This insight forms the basis of assigning a dedicated model for each tile location in our tiled ensemble method, which further extends this into a generic extension for any anomaly detection architecture.

The reduction of memory use in anomaly detection is a common research topic~\cite{batzner_efficientad, lee_memory, gudovskiy_cflow}. This is especially relevant for fields such as pathology~\cite{srinidhi_deep_histopathology}, where a high input resolution is needed to distinguish the anomalous characteristics within the images~\cite{nejat_pathoAD}. Processing images at a higher resolution prevents missing small anomalies but at the same time leads to increased GPU memory usage. In light of these challenges, the tiled ensemble extends the capabilities of various anomaly detection methods to enable efficient processing of high-resolution images. This approach capitalizes on findings from \citeauthor{heckler_backbones}~\cite{heckler_backbones}, which explores how image resolution impacts the performance of anomaly detection architectures.

The ensemble approach is often used to increase the performance of a base model. Several individual models are combined, which results in a better generalization performance~\cite{ganaie_ensemble}, accuracy, stability, and reproducibility~\cite{cao_ensemble_bio}. Ensembles are increasingly popular in visual anomaly detection and have shown promising results~\cite{shebuti_ensAD, hu_ens_rnd_proj, zahid_ibaggedfcnet, singh_ens_convnets}. \citeauthor{bergmann_uninformed}~\cite{bergmann_uninformed} used an ensemble of students to mimic the teacher network. Recent anomaly detection methods employ ensembles by keeping the architecture consistent while using a different backbone for feature extraction~\cite{bae_pni, roth_patchcore, hyun_reconpatch}. In each of these studies ensemble models consistently outperform single models, achieving state-of-the-art results in anomaly detection.

%% file: sec/3_method.tex
\section{Method}
\label{sec:method}

\begin{figure*}[t]
    \centering
    \includegraphics[width=1\linewidth]{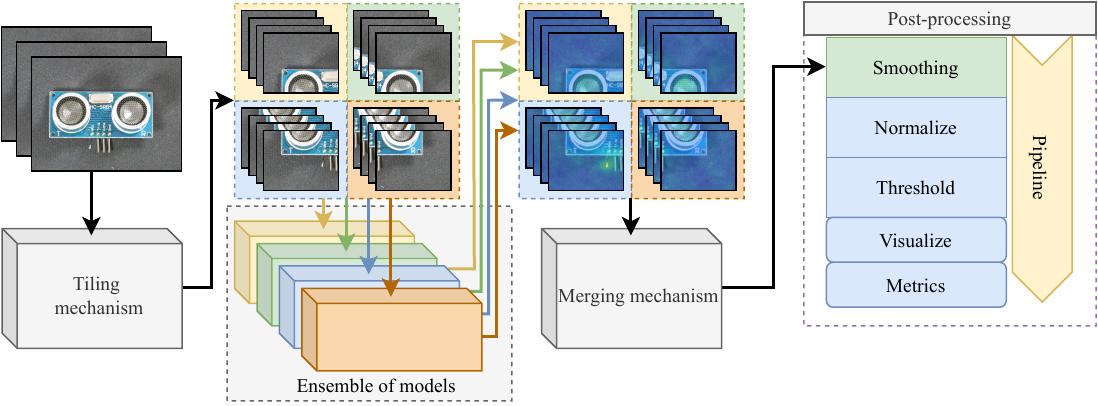}
    \caption{High-level tiled ensemble workflow: Images are first divided into tiles, and separate models are trained for each tile location. Predictions are generated individually for each tile, merged back together, and finally post-processed. Note that tiles can also be overlapping, yet the training is independent for each model.}
    \label{fig:flow}
\end{figure*}

The tiled ensemble method is structured as a series of sequential steps. The method initially divides the high-resolution image into tiles (\Cref{m:tiling}), followed by training individual models for each tile location (\Cref{m:train}). Once predictions are obtained, a merging mechanism is utilized to produce full-image-level data (\Cref{m:merge}), followed by standard post-processing steps.

A high-level overview of the workflow is presented in \Cref{fig:flow}. The approach encapsulates all the training and inference steps in a pipeline, which enables immediate application for industrial use cases that involve the analysis of high-resolution images.

\subsection{Tiling}
\label{m:tiling}

To reduce the memory footprint of a high-resolution image, the first step is to split the image into a set of tiles, which are then separately processed by an individual model.
For an image $X \in \mathbb{R}^{c \times h \times w}$, tile size $h^t \times w^t$ and stride $s_h, s_w$, this set of tiles $\mathcal{T}$ is defined as 
\begin{equation}
    \begin{aligned}
        \mathcal{T} = \{&T_{i, j} \in \mathbb{R}^{c \times h^t \times w^t} | \\
        & i \in [0,...,\left\lfloor \frac{h - h^t}{s_h} \right\rfloor], h^t \leq h, s_h \leq h\\
        & j \in [0,...,\left\lfloor \frac{w - w^t}{s_w} \right\rfloor], w^t \leq w, s_w \leq w \\
        & h^t, w^t, s_h, s_w \in \mathbb{N}\}
    \end{aligned}
    \label{eq:tile_set}
\end{equation}
If the stride and tile size don't precisely match the image, the image is padded with zeros, which are later removed during untiling.
Each tile spans the following pixels of the original image:
\begin{equation}
    \begin{aligned}
        T_{i,j} = \{(a, b) | & a \in [s_h * i,...,s_h * i + h^t), \\
                             & b \in [s_w * i,...,s_w * i + w^t)\}
    \end{aligned}
    \label{eq:tile_pix}
\end{equation}
The pixels that tiles cover can also overlap in case of stride smaller than tile size, i.e. $s_h < h^t$ or $s_w < w^t$.

For instance, consider an input image with dimensions $512 \times 512$ (height $h = 512$ and width $w = 512$), a tile size of $256 \times 256$ (tile height $h^t= 256$ and tile width $w^t = 256$) and a stride of $s_h = 128$ and $s_w = 128$. This configuration yields 9 overlapping tiles, labeled $T_{0,0}$ through $T_{2,2}$. The overlapping area between tiles $T_{0,0}$ and $T_{0, 1}$ consists of pixels ${(a, b) | a \in [0,..., 256) \land b \in [128,..., 256)}$, which means it encompasses the right half of $T_{0,0}$ and the left half of $T_{0, 1}$.

Tiling the input image enables predicting images with higher resolution while allowing models to be trained on smaller inputs. This approach significantly reduces GPU memory consumption. Another benefit of tiling is that each model is only responsible for the designated tile location. This localized processing ensures that anomalies detected within a specific tile do not influence the detection results in adjacent or distant tiles, which, overall, prevents the trigger of unrelated spurious predictions across the image.

\subsection{Training and inference}
\label{m:train}

\noindent \textbf{Training}.
By splitting the image into smaller tiles, the problem of high GPU memory consumption is efficiently addressed. However, training a single model on the combined set of all tile locations could lead to a loss of positional information, with a potential negative effect on the performance of models that perform better on aligned objects.
The tiled ensemble approach addresses this by employing a separate model for each tile location, where the underlying model architecture of the individual models remains unchanged.

Formally, a separate model $M_{i, j}$ is trained for each tile location in the entire set of tiles $\mathcal{T}$, defined in \Cref{eq:tile_set}, resulting in a set of models:
\vspace{-1.4ex}
\begin{equation}
    \begin{aligned}
        \mathcal{M} = \{M_{i, j}| M_{i, j} \text{ trained on } T_{i, j}\}
    \end{aligned}
    \label{eq:models}
\end{equation}

The tiled ensemble method requires no further modifications to the training process, which is similar to that of a single model. Since each tile location is processed independently, it is possible to train the models in parallel across different devices.

\noindent \textbf{Inference}.
Once all the models for all locations are trained, the same tiling procedure is followed and each tile is processed by the corresponding model in inference. For tile $T^{test}_{i, j}$ in inference time and model $M_{i, j}$, the pixel-level anomaly map $\mathcal{A}_{i, j}$ and anomaly score $\mathrm{s}_{i, j}$ is obtained as:
\vspace{-1.7ex}
\begin{equation}
    \begin{aligned}
        \mathcal{A}_{i, j}, s_{i, j} = M_{i, j}(T^{test}_{i, j})
    \end{aligned}
    \label{eq:inference}
\end{equation}
In this case, the score $\mathrm{s}_{i, j}$ is obtained as specified by the underlying architecture. This can either be achieved as a separate process or by taking the maximum value from $\mathcal{A}_{i, j}$.

Due to the independence of predictions, the storage of each tile predictions can also be efficiently managed. By moving the tile predictions to the main memory, the GPU memory usage remains within the constraints.

\subsection{Merging}
\label{m:merge}

\begin{figure}[!h]
  \centering
   \includegraphics[width=1\linewidth]{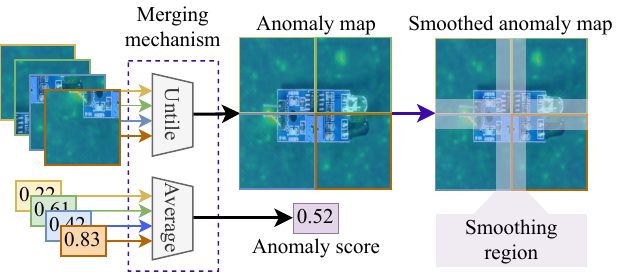}
   \caption{Overview of merging procedure and smoothing. Tile-level anomaly maps are untiled into a full image, with pixel-wise averaging applied to overlapping regions. Anomaly scores from all tiles are averaged, producing a single image-level score. After the predictions are merged, the first step of post-processing involves smoothing the region around the tile seams to enhance the quality of the anomaly map.}    
   \label{fig:joining_post}
\end{figure}

A merging mechanism is utilized to produce a full-resolution anomaly map $\mathcal{A}$ and the score $s$ from individual tile predictions $\mathcal{A}_{i, j}$ and $s_{i, j}$ (\Cref{fig:joining_post}).

Tile-level anomaly maps $\mathcal{A}_{i, j}$ are simply untiled back into a full-image representation to get $\mathcal{A}$. As the tiles can also be overlapping, a pixel-wise averaging strategy~\cite{adey_ae_wo_recon} is applied on overlapping regions.

Different strategies can be used to tackle the merging of image-level scores $s_{i, j}$. An image can be classified as anomalous as soon as one of the patches is anomalous~\cite{roth_patchcore}. Alternatively, the score over all the tiles can be averaged to obtain a single score, which is in our case the default option:
\vspace{-1.0ex}
\begin{equation}
    \begin{aligned}
        s &= \frac{1}{N} \sum_{i, j} s_{i, j}
    \end{aligned}
    \label{eq:score}
\end{equation}
The borders of the tiles create seams, leading to undesirable disturbances in the image. To mitigate this issue and enhance the outcomes, a Gaussian filter is applied for smoothing. This smoothing process is confined to a narrow region surrounding the seam, as depicted in \Cref{fig:joining_post}. The default width of the smoothing region is $10\%$ of the tile width on each side of the seam.
In line with standard anomaly detection procedures, the final classification and localization predictions can be obtained by applying a thresholding mechanism to the image-level anomaly scores and anomaly maps respectively.

%% file: sec/4_experiments.tex
\section{Experiments}
\label{sec:experiments}


An in-depth analysis of the tiled ensemble across multiple configurations and anomaly detection architectures is conducted. The following sections outline the protocols and setups employed to assess the impact of the tiled ensemble.

\subsection{Experimental details}

\noindent\textbf{Datasets.} The method is evaluated on two established industrial datasets: MVTec AD~\cite{bergmann_mvtec} and VisA~\cite{zou_visa}. MVTec AD and VisA datasets comprise 15 and 12 categories, respectively. Each category consists of a training set containing only normal images and a test set, comprising both normal and anomalous images, with their corresponding pixel-precise ground truth annotations. The anomalies vary in types, shapes, and scales, with the prevalence of larger anomalies in MVTec AD and smaller defects in VisA. An analysis of defect scales is presented in \Cref{ap:ano_size}.

For both datasets, the images are of high resolution and are resized according to the specified dimensions in the experimental setups, as detailed in the following sections.

\noindent\textbf{Evaluation Metrics} Both image and pixel-level performance are evaluated using standard anomaly detection metrics. For image-level anomaly detection, the Area Under the Receiver Operator Curve (AUROC) is employed. To evaluate pixel-wise performance in anomaly localization, the Area Under the Per-Region-Overlap Curve (AUPRO) is used.

The exact protocol outlined by the authors of EfficientAD~\cite{batzner_efficientad} is followed to obtain latency, throughput, and inference GPU memory consumption. The only exception is the usage of a batch size of 8 instead of 16 for PatchCore, due to excessive GPU memory usage in the case of a single model with $512\times512$ resolution. For a tiled ensemble, the benchmark inference step encapsulates tiling, inference on all tiles, and untiling. Experiments were conducted on a system with Intel(R) Xeon(R) Gold 5320 CPU and Nvidia Tesla A100 GPU (training) and Nvidia Tesla V100S (inference).

\subsection{Evaluation setups}
\label{exp:setups}

To comprehensively evaluate and compare the performance of the tiled ensemble, four architectures from diverse paradigms are employed. Padim~\cite{defard_padim} covers probability density modelling, Patchcore~\cite{roth_patchcore} is a memory bank based approach, Reverse Distillation~\cite{deng_rev_dist} represents student-teacher architectures, and Fastflow~\cite{yu_fastflow} normalizing flows. Each architecture is then trained in six different setups, where two use a single model with varying resolution, two employ tiled input to a single model and the remaining two utilize the tiled ensemble.
\vspace{1ex}

\noindent\textbf{Single model with 256px image size -- \textit{SM256}}. A single model for each architecture is trained with an input size of $256\times256$ pixels, aligning with the tile size of the ensemble models. While the effective final resolution processed by this setup is smaller, this setup serves as a baseline as this resolution is the most common in other works.
\vspace{1ex}

\noindent\textbf{Single model with 512px image size -- \textit{SM512}}. To explore the effect of resolution without ensembling, a single model is trained with an input image size of $512\times512$ pixels. In this case, the model processes the same effective resolution as our base tiled ensemble. However, it consumes a larger amount of GPU memory. This setup allows for a comparison of memory usage and the extent to which the benefits result from ensembling rather than increased resolution.
\vspace{1ex}

\noindent\textbf{Tiled ensemble with 9 overlapping 256px tiles -- \textit{ENS9}}.
The base tiled ensemble setup has image resolution of $512\times512$ pixels, which is then divided into 9 overlapping $256\times256$ tiles ($h^t=w^t=256$, $s_h=s_w=128$). The final predicted anomaly map maintains the same dimensions as the input image, i.e. $512\times512$. This setup is utilized to highlight the effects of ensembling properties in addition to the ability to process high resolution while adhering to memory constraints.
\vspace{1ex}

\noindent\textbf{Tiled ensemble with 4 non-overlapping 256px tiles -- \textit{ENS4}}.
In this setup, the input image has a resolution of $512\times512$ and is divided into four non-overlapping $256\times256$ tiles ($h^t=w^t=256$, $s_h=s_w=256$). The dimension of the predicted anomaly map remains consistent with the input image, i.e. $512\times512$s. This setup is utilized to highlight the efficient processing of high resolution, without the additional benefits of multiple (overlapping) predictions as in \textit{ENS9}.
\vspace{1ex}

\noindent\textbf{Single model with 9 overlapping 256px tiles -- \textit{ST9}}. This setup involves using a single model trained on tiled input. The image resolution remains at $512\times512$ pixels, divided into nine overlapping $256\times256$ tiles ($h^t=w^t=256$, $s_h=s_w=128$), and stacked batch-wise. A single model in this case receives a $256\times256$ tile for input. This setup is used to compare the effect of having a separate model in a tiled ensemble specializing solely in a single tile location. 
\vspace{1ex}

\noindent\textbf{Single model with 512px with 4 non-overlapping 256px tiles -- \textit{ST4}}. Matching the tiled ensemble setup without overlapping tiles, this setup explores how tiling the input works in the case of utilizing a single model for all tile locations. Here, a $512\times512$ input image is split into four non-overlapping $256\times256$ tiles ($h^t=w^t=256$, $s_h=s_w=256$), and stacked batch-wise. A single model is then trained on these tiles, with an input size of $256\times256$ pixels.\\

\noindent\textbf{Common properties}. Each setup is trained on every category, with every run repeated 3 times using a different random seed. A consistent batch size of 32 is used, except for Patchcore where a batch size of 8 is used due to memory limitations. The backbone used in all setups is ResNet18, to keep the comparison fair. Following~\cite{batzner_efficientad, heckler_backbones} FastFlow, and Reverse Distillation are limited to 200 steps for all setups. Other properties are kept the same as provided by the original authors of the models and as implemented in Anomalib~\cite{anomalib}. 

%% file: sec/5_results.tex
\section{Results}
\label{sec:res}

\noindent\textbf{Results on MVTec AD.} \Cref{tab:mvtec} reports anomaly detection and localization results obtained on the MVTec AD dataset. A tiled ensemble with overlapping tiles (ENS9) achieves the best results in terms of anomaly detection for Padim and FastFlow and second best for PatchCore and Reverse Distillation. It also achieves the best localization performance for PatchCore and FastFlow, and second best for Padim.
\begin{table}[!h]
    \centering
    \setlength{\tabcolsep}{1pt}
    \begin{tabular}{l||c|c|c|c}
    Setup& PatchCore& Padim& FastFlow& \makecell{Reverse \\ Distillation}\\ \hline
    SM256& 97.7/92.8& \bm2{89.2}/\bm1{91.2}& \bm2{93.1}/89.1& \bm1{90.8}/\bm1{89.5}\\
    SM512& \bm1{98.0}/\bm2{94.5}& 83.0/\bm2{91.0}& 90.5/88.5& 78.5/\bm2{87.2}\\
    ST4& \bm2{97.8}/94.0& 83.8/90.6& 91.4/85.0& 80.7/86.2\\
    ST9& \bm2{97.8}/94.3& 83.3/90.6& 90.1/82.8& 77.3/\bm1{89.5}\\
    ENS4& 96.5/94.1& 87.3/90.5& 91.8/\bm2{89.5}& 84.5/82.6\\
    ENS9& \bm2{97.8}/\bm1{95.3}& \bm1{89.7}/\bm2{91.0}& \bm1{95.0}/\bm1{91.4}& \bm2{87.8}/82.6\\
    \end{tabular}
    \caption{Results in anomaly detection and localization (AUROC/AUPRO) on MVTec AD. \textcolor{goldD}{Best} and \textcolor{silverD}{second best} results are marked. A mean of 3 runs is reported for each setup.}
    \label{tab:mvtec}
\end{table}

Results on VisA with all 6 setups are displayed in~\Cref{tab:visa}. A tiled ensemble with overlapping tiles (ENS9) achieves the best anomaly detection results for Padim, FastFlow, and Reverse Distillation, significantly outperforming baseline single model (SM256) and single model processing the same resolution of $512 \times 512$ (SM512) in all three cases. ENS9 also achieves the best anomaly localization results for FastFlow and Reverse Distillation.\\

\begin{table}[!ht]
    \centering
    \setlength{\tabcolsep}{1pt}
        \begin{tabular}{l||c|c|c|c}
		 Setup& PatchCore& Padim& FastFlow& \makecell{Reverse \\ Distillation}\\ \hline
SM256& 92.0/87.7& 83.7/82.1& 87.4/81.4& 83.2/86.9\\
SM512& \bm1{97.2}/\bm1{94.0}& 81.9/86.9& \bm2{89.8}/\bm2{88.4}& 71.5/89.3\\
ST4& 95.2/93.6& 81.8/\bm1{87.2}& 87.1/83.4& 72.5/89.3\\
ST9& \bm2{96.7}/\bm2{93.7}& 82.3/\bm2{87.1}& 85.6/78.6& 80.6/88.9\\
ENS4& 93.1/93.0& \bm2{83.8}/86.9& 89.4/86.8& \bm2{88.3}/\bm2{89.6}\\
ENS9& 95.4/\bm2{93.7}& \bm1{86.3}/86.9& \bm1{92.5}/\bm1{89.2}& \bm1{91.4}/\bm1{90.0}\\
\end{tabular}
    \caption{Results in anomaly detection and localization (AUROC/AUPRO) on VisA. \textcolor{goldD}{Best} and \textcolor{silverD}{second best} results are marked. A mean of 3 runs is reported for each setup.}
    \label{tab:visa}
\end{table}
\begin{figure*}[!ht]
  \centering
   \includegraphics[width=1\linewidth]{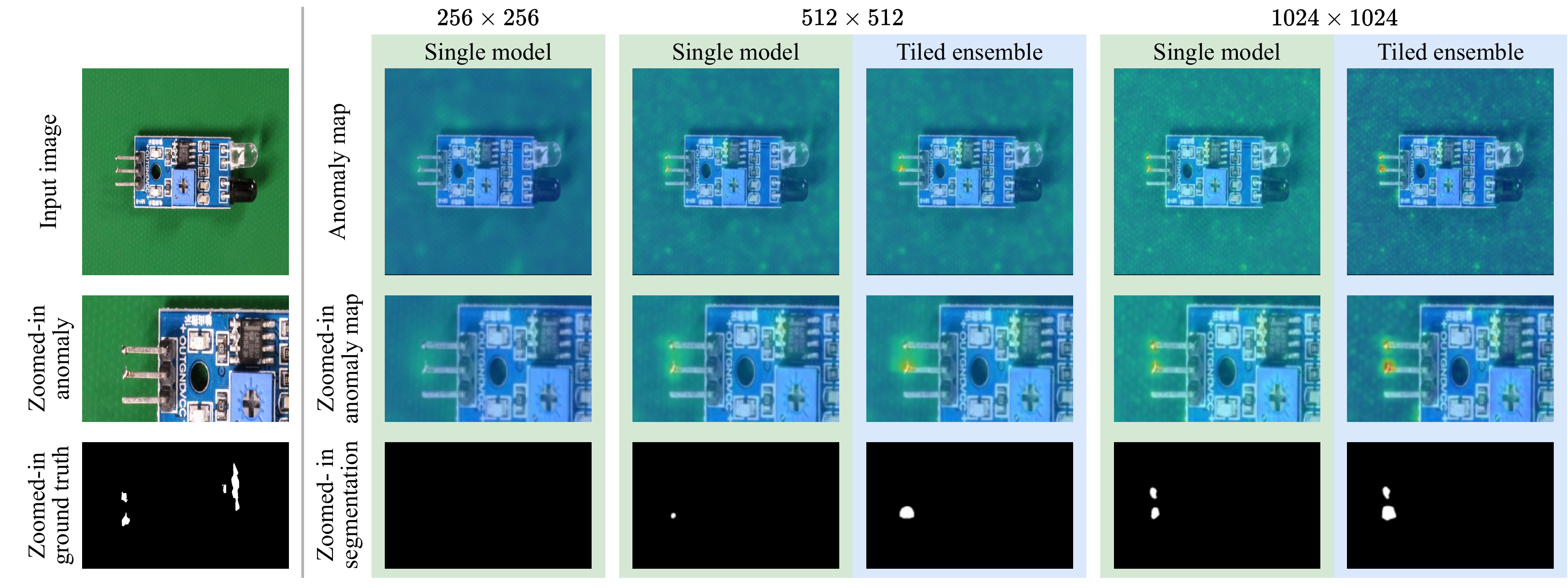}
   \caption{Effects of increasing resolution for Padim on VisA PCB3 category. The first column contains full and zoomed-in input images with corresponding ground truth. The next columns depict full and zoomed-in anomaly maps with their corresponding binary segmentation. The resolution of results is written above each block. Notice how localization improves with higher resolution. The tiled ensemble uses a setup with $h^t=w^t=256$, $s_h=s_w=256$.}
   \label{fig:resolution}
\end{figure*}

The anomaly detection results of a tiled ensemble (ENS4 and ENS9) consistently outperform a single model with tiled input (ST4 and ST9) in almost all setups, except for PatchCore, and in terms of localization for Padim. This indicates that having a separate model specialized in each tile location can lead to better performance in high-resolution images. The tiled ensemble with overlapping tiles (ENS9) in most setups outperforms a single model processing the same resolution (SM512) as well as a tiled ensemble with non-overlapping tiles (ENS4). This demonstrates the potential benefits of the stacking ensemble mechanism. 

In the tiled ensemble setup, the kNN search in PatchCore's memory bank is limited to embeddings from within the same tile location, whereas single-model setups provide access to embeddings from the entire image. This may explain why PatchCore tends to benefit from a single-model setup.
In both MVTec AD and VisA, Reverse Distillation and Padim struggle if the resolution is increased without utilizing the tiled ensemble, showing subpar performance when comparing SM512 to baseline SM256. In the case of MVTec AD, Reverse Distillation still works best with the baseline model, indicating that for some architectures and large anomalies, a tiled ensemble is not necessarily needed.

More detailed results for each category and setup with included standard deviation on MVTec AD and VisA are included in~\Cref{ap:cat_res}. 

\begin{figure}[!ht]
  \centering
   \includegraphics[width=1\linewidth]{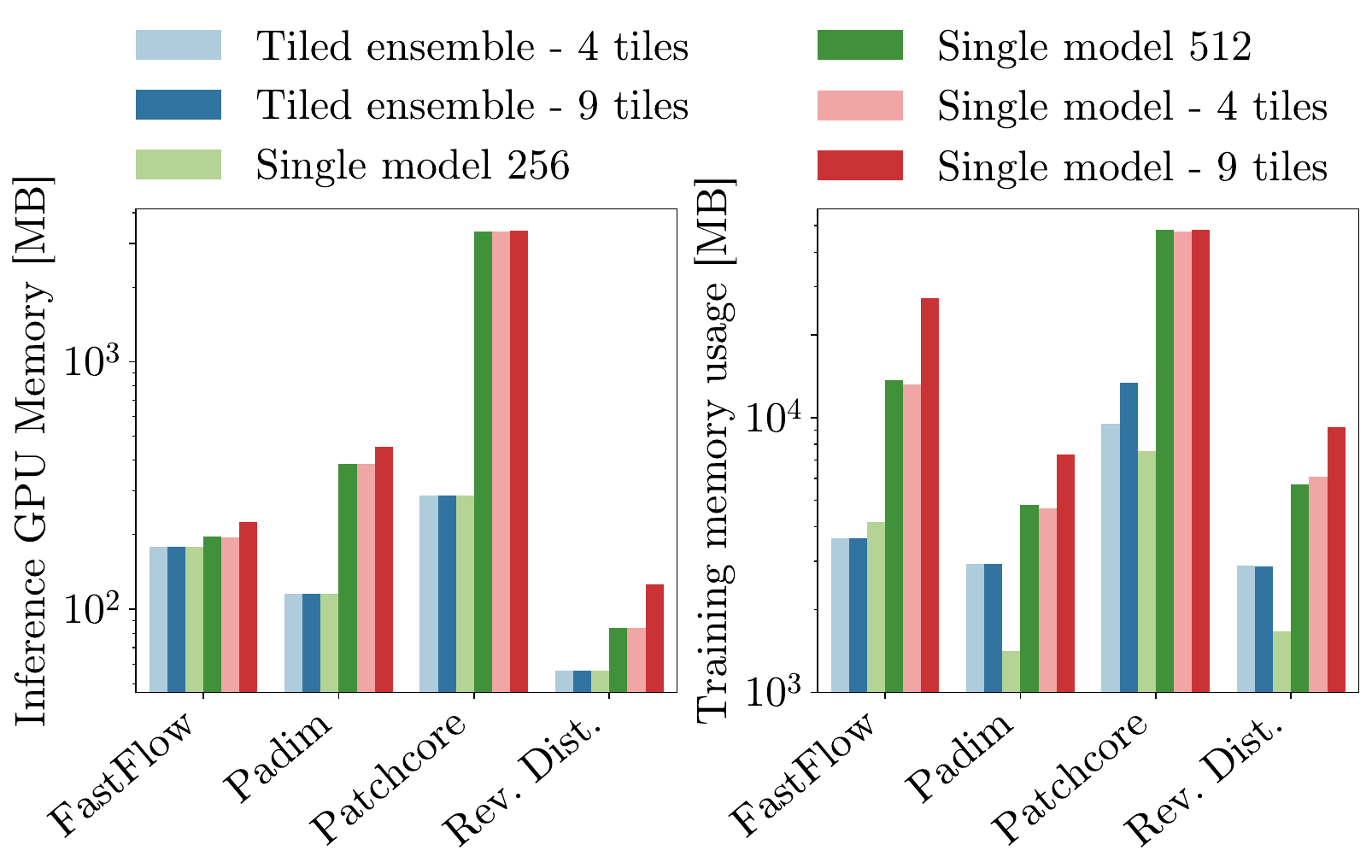}
   \caption{Inference and training GPU memory usage. The memory consumption of the tiled ensemble remains within the range of a model processing an image with the resolution of a single tile (\textit{Single model 256}). This for some models results in a notable reduction, particularly evident in models such as Patchcore.}    
   \label{fig:memory}
\end{figure}

\noindent\textbf{GPU memory usage.} Inference GPU memory and training GPU memory usage are presented in~\Cref{fig:memory}, respectively. 
The tiled ensemble is unaffected by the number of tiles as long as the tiles have the same resolution. During inference, memory consumption remains comparable to that of a single model processing the resolution equivalent to a tile (\textit{Single model 256}), across all models.

Inference GPU memory consumption holds significant importance for end-applications, but training memory consumption also poses a challenge with larger image resolutions. The tiled ensemble maintains a manageable memory footprint in both training and inference, roughly equating to the memory consumption of a single model (\Cref{fig:memory}). Note that the relative memory advantage of the tiled ensemble further increases for higher effective image resolutions (provided that the tile size remains the same), as the memory consumption of each individual model is only related to the tile size.

\begin{figure}[!h]
  \centering
   \includegraphics[width=1\linewidth]{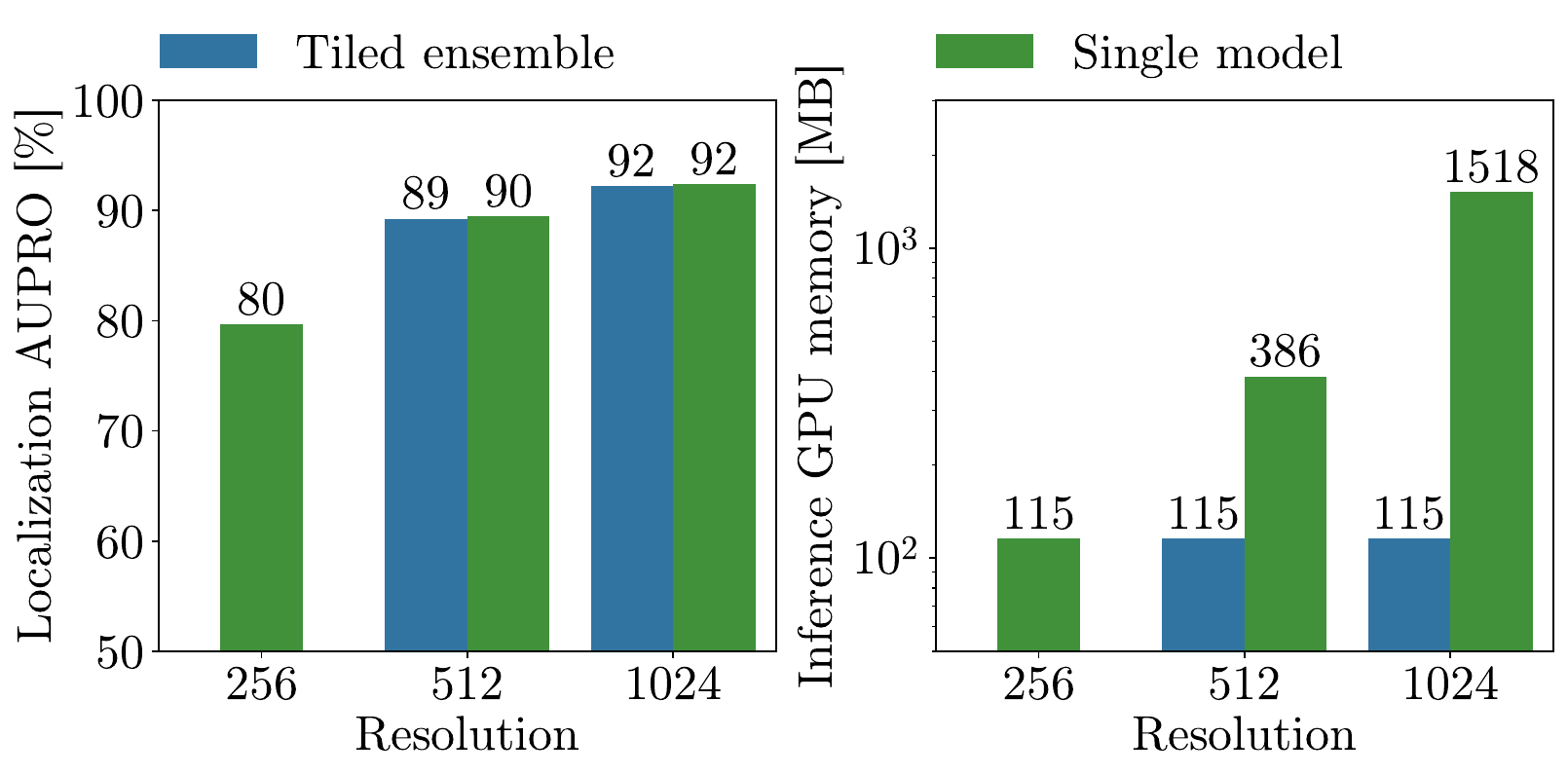}
   \caption{Localization results in terms of AUPRO on VisA PCB3 category for a single model and tiled ensemble with different resolutions. The corresponding memory consumption of each setup is shown on the right. The tiled ensemble uses a setup with $h^t=w^t=256$, $s_h=s_w=256$.}    
   \label{fig:resolution_quant}
\end{figure}

\noindent\textbf{Effect of resolution on performance and GPU memory usage.} A case study is performed on the PCB3 category of the VisA dataset, which contains many small defects that can benefit from increased resolution. The Padim model is used to explore memory consumption and verify the effect of resolution on localization performance. A tile size of 256 with stride 256 is used for ensemble setup ($h^t=w^t=256$, $s_h=s_w=256$). The results of localization performance with respect to resolution are presented in~\Cref{fig:resolution_quant} with GPU memory consumption also reported for each setup.

Increased resolution offers increased localization performance, most notably showing an improvement in the initial increase from $256 \times 256$ to $512 \times 512$, at which many small anomalies already become better detectable. While the memory of a single model processing a larger resolution steeply increases, the tiled ensemble maintains the same memory consumption for all resolutions. \cref{fig:resolution} contains a qualitative example depicting the localization of a small anomaly from this experiment.

\begin{figure}[!t]
  \centering
   \includegraphics[width=1\linewidth]{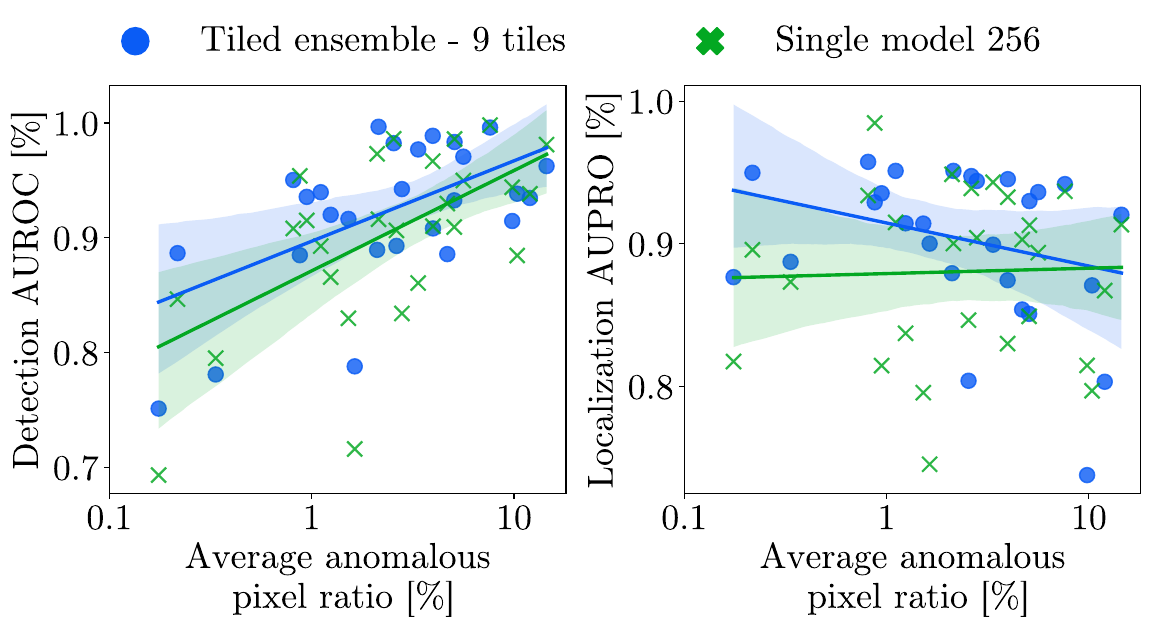}
   \caption{
   Effect of defect size on anomaly detection and localization performance of the tiled ensemble (ENS9) and single model with $256 \times 256$ resolution (SM256). Each point represents a single dataset category from MVTec AD or VisA. Trend lines and confidence intervals added for interpretability. X-axis: average number of anomalous pixels per defective image relative to image size. Y-axis: average performance of setup across all four model architectures.
   }    
   \label{fig:size_trend}
\end{figure}

\noindent\textbf{Effect of input resolution on small anomaly detection.}
\Cref{fig:size_trend} illustrates how small anomaly detection may benefit from the higher input resolutions that can be achieved by the tiled ensemble approach. Compared to the $256 \times 256$ baseline, the tiled ensemble achieves a notable boost in both detection and localization performance for datasets in which the average size of the anomalous regions is small. As the average size of the anomalies increases, the effect diminishes and the performance of both setups converges. By increasing the effective input resolution, the tiled ensemble approach facilitates the detection and localization of small anomalous regions that would otherwise go unnoticed as a result of downscaling the input images.

\begin{figure}[!t]
  \centering
   \includegraphics[width=1\linewidth]{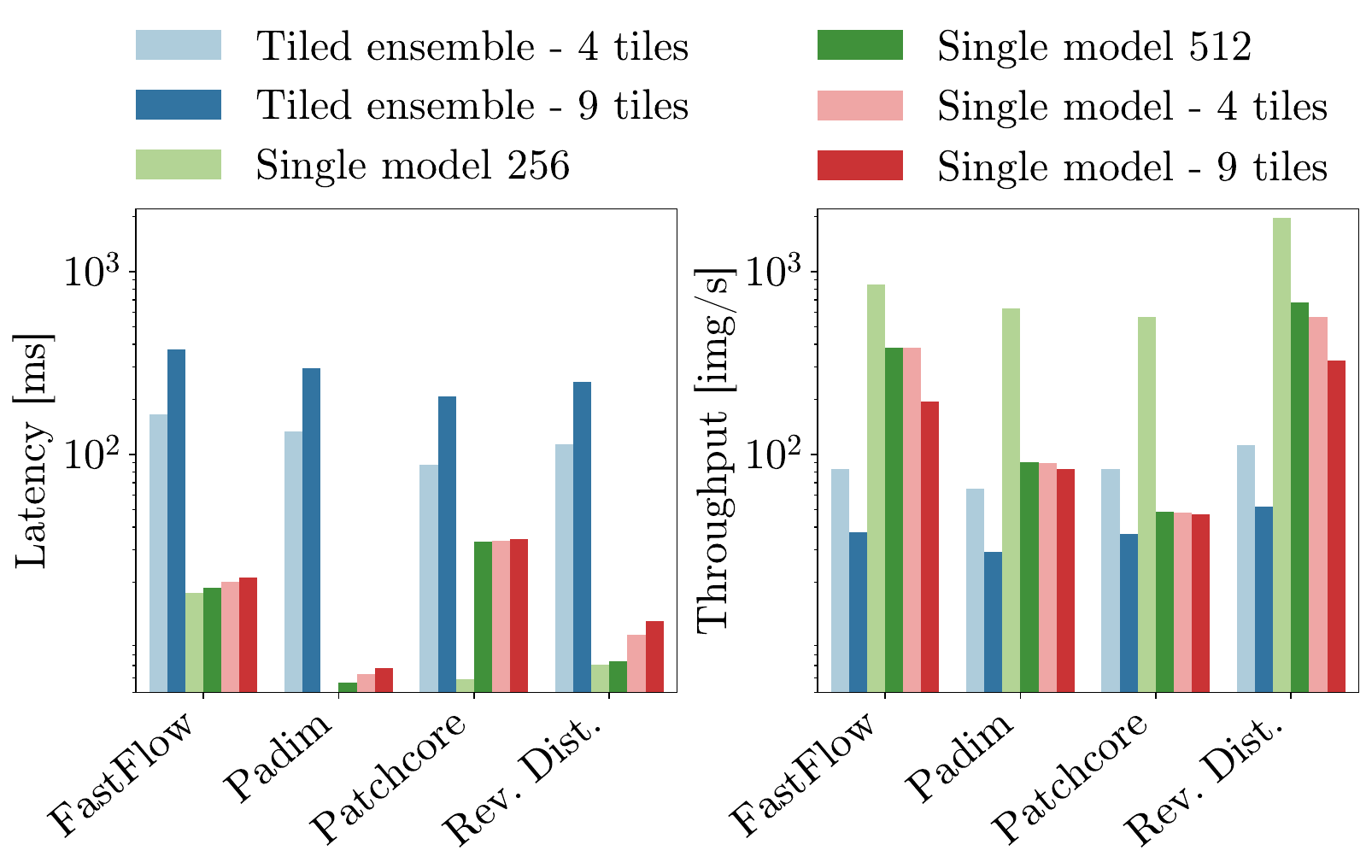}
   \caption{Latency and throughput measured for each setup on an Nvidia Tesla V100S. While the latency considerably increases, throughput sees a relatively smaller reduction for tiled ensemble configurations.}    
   \label{fig:timing}
\end{figure}

\noindent\textbf{Latency and throughput.}
The latency and throughput are presented in~\Cref{fig:timing}.
The low throughput and high latency of ENS4 and ENS9 can be attributed to the increased computational complexity of these setups (\Cref{ap:param}) and show that the performance advantage of the tiled ensemble comes at the cost of an increased runtime. The latency overhead likely stems from the time needed to transfer individual models to GPU and back, which does not affect throughput as significantly since the model's time on GPU is better utilized. For some models like Patchcore, the throughput of a 4-tiled ensemble exceeds that of a single model processing an equivalent resolution.
The preliminary studies showed that the time needed for tiled ensemble inference on GPU still outperforms the inference of a single model with increased resolution on CPU in terms of latency by around 4 times, and in terms of throughput by around 80 times. The training time of all setups is presented in \Cref{ap:time}.

%% file: sec/6_conclusion.tex
\section{Conclusion}
\label{sec:conclusion}


This paper introduces a tiled ensemble approach to effectively detect and localize small anomalies in high-resolution images, which has been a challenge due to the high GPU memory demands required by existing approaches. The tiled ensemble approach addresses this by dividing the image into smaller tiles and training a dedicated model for each tile location. This strategy ensures that the GPU memory usage remains comparable to that of a single model that processes an image the size of one tile. By employing overlapping tiles, the tiled ensemble also takes advantage of the performance improvements associated with traditional stacking ensemble methods, which further improve performance compared to those achievable by simply increasing image resolution.

The tiled ensemble is designed to be easily integrated into current anomaly detection architectures without necessitating any architectural changes, which makes it a flexible and practical solution for small anomaly detection within high-resolution imagery. 
In an extensive evaluation using various model architectures and two established datasets, the method demonstrated notable performance improvement compared to setups processing images at a lower resolution or without employing a tiled ensemble, with a particularly pronounced impact on datasets with small anomalies.

The results presented in this paper demonstrate the feasibility of applying existing or next-generation anomaly detection models within high-resolution imagery, which opens up new possibilities across various industries.

\noindent\textbf{Limitations.} Despite its promising statistical results, our approach has a notable latency overhead, which can be partially mitigated through batched inference. This can be a reasonable sacrifice in cases where the resolution is very large, to enable detection with resolutions that previously were not feasible. As suggested by \citeauthor{heckler_backbones}~\cite{heckler_backbones} and verified by \citeauthor{heckler_AD_featSel}~\cite{heckler_AD_featSel}, strategically choosing a single layer can outperform an ensemble of multiple layers and backbones in certain scenarios. Building on this insight, future research should investigate whether selecting the most suitable layer, or set of layers, for each tile location could further optimize anomaly detection in high-resolution images.
Finally, the experiments of the current study did not cover logical anomaly detection benchmarks, which could potentially suffer from a loss of global context as a result of processing each tile location separately.

%% file: sec/X_suppl.tex
\clearpage
\setcounter{page}{1}
\maketitlesupplementary

\appendix
\section{Anomaly scales per category}
\label{ap:ano_size}

\begin{figure}[!h]
    \centering
    \includegraphics[width=1\linewidth]{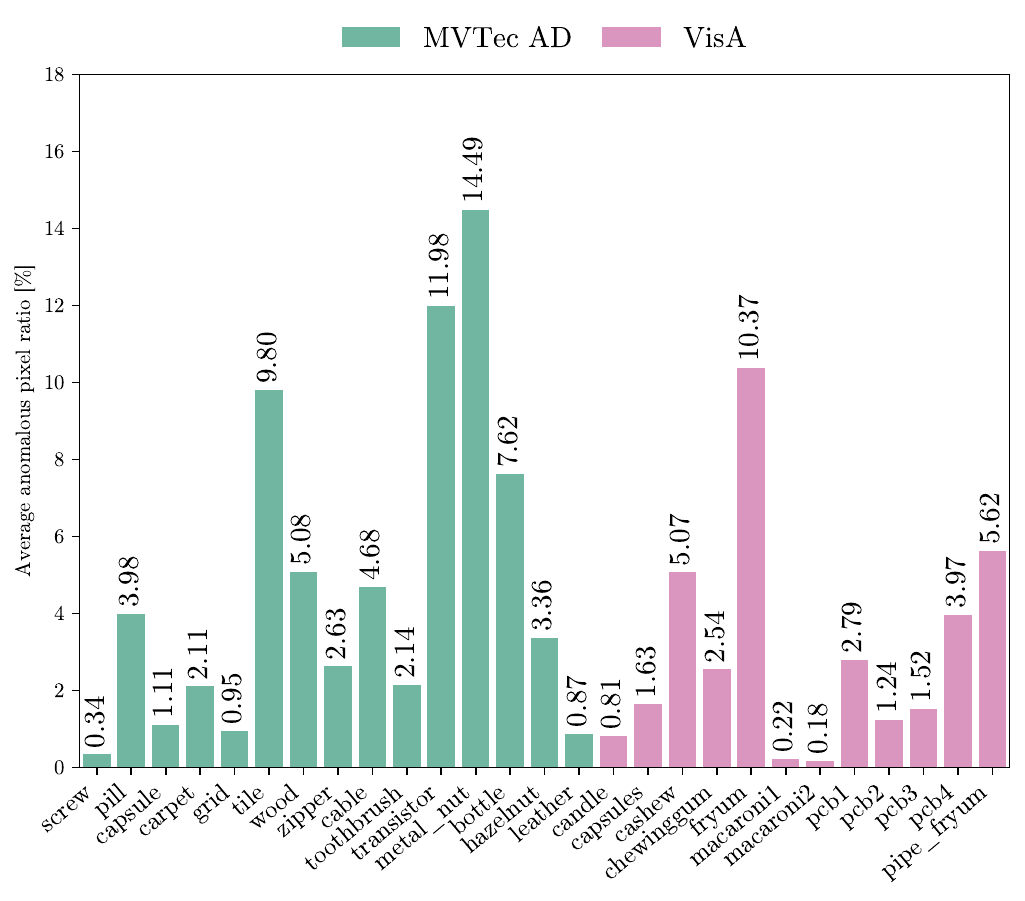}
    \caption{Average anomalous pixel ratio of defective images per category for all categories present in VisA and MVTec AD.}
    \label{fig:ano_size}
\end{figure}

\Cref{fig:ano_size} shows the average anomalous pixel ratio (what percentage of defective image is covered by the defect) present in the test set of each category from MVTecAD~\cite{bergmann_mvtec} and VisA~\cite{zou_visa}. The ratio is calculated on all defective images from the category with a resolution of $512\times512$. VisA categories contain notably smaller defects, especially in categories such as candle, macaroni 1, and macaroni 2, where the anomalous pixels on average cover less than $1 \%$ of the image.

\section{Parameter sizes}
\label{ap:param}

\Cref{tab:params} contains the parameter count (in millions) for each setup and architecture. Since the tiled ensemble consists of unchanged underlying architectures, the parameter sizes are increased by the factor equivalent to the number of models in an ensemble.

\begin{table}[!ht]
    \centering
    \setlength{\tabcolsep}{3pt}
    \begin{tabular}{l|cccc}
		  Setup& PatchCore& Padim& FastFlow& \makecell{Reverse \\ Distillation}\\ \hline
        SM256& 2.8& 2.8& 9.7& 18.2\\
        SM512& 2.8& 2.8& 12.5& 18.2\\
        ST4& 2.8& 2.8& 9.7& 18.2\\
        ST9& 2.8& 2.8& 9.7& 18.2\\
        ENS4& 11.1& 11.1& 38.9& 72.7\\
        ENS9& 25.0& 25.0& 87.6& 163.5\\
    \end{tabular}
        \caption{Parameters (\textbf{million)} for each architecture and each setup.}
    \label{tab:params}
\end{table}

\section{Training time}
\label{ap:time}

\Cref{tab:train_time} contains the training duration of each architecture with each setup. The results are averaged across all categories from MVTecAD~\cite{bergmann_mvtec} and VisA~\cite{zou_visa} and 3 runs with different seeds. In cases of FastFlow and Reverse Distillation, the training time is approximately extended by a factor equivalent to the number of models in an ensemble. This is expected due to the cumulative increase in epochs, while the training workload of each model inside the ensemble remains in line with a single model processing a $256 \times 256$ resolution.

In the case of PatchCore and Padim, this does not hold since they are not trained using backpropagation. Therefore, the time required for training doesn't scale equivalently when directly increasing resolution (from SM256 to SM512) or when achieving higher processed resolution through multiple smaller models operating within an ensemble (from SM256 to ENS4/ENS9).

\begin{table}[!ht]
    \centering
    \setlength{\tabcolsep}{3pt}
    \begin{tabular}{l|cccc}
		  Setup& PatchCore& Padim& FastFlow& \makecell{Reverse \\ Distillation}\\ \hline
        SM256& 213.1& 772.0& 556.3& 721.5\\
        SM512& 2273.0& 2423.5& 799.6& 838.8\\
        ST4& 2254.6& 2400.7& 805.1& 836.5\\
        ST9& 2275.2& 2423.0& 851.9& 867.5\\
        ENS4& 879.9& 444.9& 2561.8& 2970.9\\
        ENS9& 1810.6& 974.6& 5857.7& 6876.8\\
    \end{tabular}
    \caption{Training time in \textbf{seconds} for each architecture and each setup. Results are averaged over all categories and 3 runs with different seeds.}
    \label{tab:train_time}
\end{table}

\onecolumn
\section{Results of all categories for each setup}
\label{ap:cat_res}

This section contains tables with category-specific results for all architectures and their setups.
\begin{table*}[!h]
    \centering
    \setlength{\tabcolsep}{1pt}
    \begin{tabular}{lcccccc}
        \toprule
		~  & \makecell{Single model \\ 256} & \makecell{Single model \\  512} & \makecell{Tiled ensemble \\ 4 tiles} & \makecell{Tiled ensemble \\ 9 tiles} & \makecell{Single model \\ 4 tiles} & \makecell{Single model \\ 9 tiles}\\ \hline
		Carpet & 97.4/94.3& 96.5/97.0& 95.2/96.0& 97.6/\textbf{97.0}& \textbf{98.1}/96.2& 95.7/96.0\\
		Grid & 96.3/88.8& 99.9/\textbf{97.1}& 96.4/96.1& 96.9/96.7& \textbf{100.0}/96.8& \textbf{100.0}/96.8\\
		Leather & \textbf{100.0}/97.3& 99.4/98.8& 98.0/98.4& 98.8/\textbf{98.9}& 99.7/98.4& 99.1/98.6\\
		Tile & \textbf{99.5}/83.9& 97.8/88.0& 94.9/86.1& 96.3/\textbf{88.0}& 97.1/86.0& 96.4/86.5\\
		Wood & 99.2/89.1& 99.1/94.2& 99.5/93.7& \textbf{99.7}/\textbf{94.3}& 99.1/93.9& 99.0/94.0\\\midrule
		Bottle & \textbf{100.0}/93.3& \textbf{100.0}/96.3& \textbf{100.0}/96.0& \textbf{100.0}/\textbf{96.4}& \textbf{100.0}/96.0& \textbf{100.0}/96.1\\
		Cable & \textbf{98.0}/93.9& 95.6/93.4& 95.0/93.5& 96.9/\textbf{94.6}& 94.2/92.7& 96.3/93.0\\
		Capsule & 98.7/93.1& 99.1/96.9& 99.5/96.2& \textbf{99.8}/96.9& 99.3/96.4& 99.7/\textbf{97.0}\\
		Hazelnut & 100.0/95.0& \textbf{100.0}/\textbf{97.0}& 99.7/96.3& 99.8/97.0& \textbf{100.0}/96.2& \textbf{100.0}/96.7\\
		Metal nut & 99.8/94.2& \textbf{100.0}/\textbf{96.1}& 97.4/95.1& 98.7/95.7& 99.7/95.4& 99.9/95.5\\
		Pill & \textbf{94.0}/93.7& 92.3/\textbf{96.5}& 91.0/95.7& 92.8/96.4& 90.2/95.8& 93.5/96.5\\
		Screw & 93.7/96.6& \textbf{98.3}/97.8& 84.6/97.8& 92.1/\textbf{98.2}& 97.0/97.8& 98.1/98.1\\
		Toothbrush & 94.4/90.9& 96.2/\textbf{96.0}& 98.9/95.8& \textbf{99.7}/95.9& 96.7/95.6& 95.1/95.7\\
		Transistor & 98.0/\textbf{92.6}& 97.2/74.6& 97.2/77.6& \textbf{98.4}/86.1& 97.4/75.1& 96.8/76.6\\
		Zipper & 95.9/95.1& 98.6/\textbf{97.8}& \textbf{99.5}/96.9& 99.0/97.6& 98.7/97.1& 97.0/97.0\\ \midrule
		\textit{Average} & \makecell{ 97.7/92.8 \\ ($\pm0.06$ / $\pm 0.04$)}& \makecell{ \textbf{98.0}/94.5 \\ ($\pm0.08$ / $\pm 0.01$)}& \makecell{ 96.5/94.1 \\ ($\pm0.03$ / $\pm 0.04$)}& \makecell{ 97.8/\textbf{95.3} \\ ($\pm0.04$ / $\pm 0.01$)}& \makecell{ 97.8/94.0 \\ ($\pm0.03$ / $\pm 0.01$)}& \makecell{ 97.8/94.3 \\ ($\pm0.37$ / $\pm 0.05$)}\\
      \bottomrule
    \end{tabular}
    \caption{MVTec AD results of all 6 setups for Patchcore. The row contains results for a particular category, with columns containing detection and localization results (AUROC/AUPRO) for each setup. The mean of 3 runs is reported with the corresponding standard deviation in parentheses. The best result for each category is in \textbf{bold}.}
    \label{tab:mvtec_patchcore}
\end{table*}

\begin{table*}[!h]
    \centering
        \setlength{\tabcolsep}{1pt}
    \begin{tabular}{lcccccc}
    \toprule
		~  & \makecell{Single model \\ 256} & \makecell{Single model \\  512} & \makecell{Tiled ensemble \\ 4 tiles} & \makecell{Tiled ensemble \\ 9 tiles} & \makecell{Single model \\ 4 tiles} & \makecell{Single model \\ 9 tiles}\\ \hline
		Carpet & \textbf{96.7}/\textbf{95.3}& 92.3/94.0& 89.2/93.4& 93.7/94.2& 92.4/93.5& 91.6/93.3\\
		Grid & 86.4/79.5& \textbf{93.5}/87.2& 87.6/86.5& 89.3/\textbf{87.5}& 92.8/86.6& 92.7/86.9\\
		Leather & \textbf{98.0}/93.6& 96.1/97.9& 95.0/98.0& 96.6/\textbf{98.2}& 96.4/98.0& 96.0/97.9\\
		Tile & \textbf{94.8}/\textbf{82.0}& 84.4/73.2& 89.8/73.2& 90.2/73.8& 84.5/73.2& 84.4/72.5\\
		Wood & \textbf{98.1}/92.6& 94.8/94.1& 96.4/94.0& 97.1/\textbf{94.3}& 94.3/94.0& 94.8/94.1\\\midrule
		Bottle & 99.5/95.2& 98.8/95.4& 99.1/95.9& \textbf{99.9}/95.7& 98.7/\textbf{95.9}& 98.8/95.5\\
		Cable & \textbf{82.3}/\textbf{89.2}& 74.1/81.3& 77.0/82.4& 81.7/82.8& 76.6/82.4& 74.8/82.2\\
		Capsule & 84.3/92.9& 81.5/93.9& 88.2/93.8& \textbf{88.8}/\textbf{93.9}& 81.2/93.9& 82.0/93.8\\
		Hazelnut & 80.0/94.3& 69.4/95.8& 87.6/95.8& \textbf{94.9}/95.9& 71.9/95.8& 70.1/\textbf{95.9}\\
		Metal nut & \textbf{96.4}/\textbf{92.2}& 93.4/89.9& 90.6/90.0& 93.8/90.5& 93.8/90.0& 93.2/90.0\\
		Pill & \textbf{86.8}/\textbf{94.4}& 72.9/93.4& 81.6/93.3& 81.4/93.6& 75.1/93.3& 73.7/94.4\\
		Screw & \textbf{74.2}/91.7& 61.3/\textbf{92.3}& 66.2/92.0& 69.4/92.2& 58.7/92.0& 61.1/92.2\\
		Toothbrush & 86.9/93.3& 88.1/95.7& 94.7/95.8& \textbf{99.4}/95.8& 89.8/\textbf{95.8}& 88.1/95.8\\
		Transistor & \textbf{89.8}/\textbf{89.3}& 78.2/87.5& 83.8/81.5& 87.6/83.2& 83.5/81.5& 81.0/82.4\\
		Zipper & \textbf{83.1}/\textbf{93.0}& 66.1/92.7& 82.9/92.7& 81.2/93.0& 67.3/92.7& 66.9/92.5\\\midrule
		\textit{Average} & \makecell{ 89.2/\textbf{91.2} \\ ($\pm1.29$ / $\pm 0.62$)}& \makecell{ 83.0/91.0 \\ ($\pm1.28$ / $\pm 0.36$)}& \makecell{ 87.3/90.5 \\ ($\pm0.76$ / $\pm 0.41$)}& \makecell{ \textbf{89.7}/91.0 \\ ($\pm0.89$ / $\pm 0.42$)}& \makecell{ 83.8/90.6 \\ ($\pm1.07$ / $\pm 0.41$)}& \makecell{ 83.3/90.6 \\ ($\pm1.12$ / $\pm 0.43$)}\\

    \bottomrule
    \end{tabular}
    \caption{MVTec AD results of all 6 setups for Padim. The row contains results for a particular category, with columns containing detection and localization results (AUROC/AUPRO) for each setup. The mean of 3 runs is reported with the corresponding standard deviation in parentheses. The best result for each category is in \textbf{bold}.}
    \label{tab:mvtec_padim}
\end{table*}

\begin{table*}[!ht]
    \centering
        \setlength{\tabcolsep}{1pt}
    \begin{tabular}{lcccccc}
    \toprule
		~  & \makecell{Single model \\ 256} & \makecell{Single model \\  512} & \makecell{Tiled ensemble \\ 4 tiles} & \makecell{Tiled ensemble \\ 9 tiles} & \makecell{Single model \\ 4 tiles} & \makecell{Single model \\ 9 tiles}\\ \hline
		Carpet & \textbf{99.0}/\textbf{93.8}& 94.8/87.7& 70.9/90.2& 82.0/91.9& 95.5/87.3& 96.3/88.3\\
		Grid & 98.8/93.2& 99.5/96.9& 96.8/95.5& 97.1/\textbf{96.9}& \textbf{99.9}/95.8& 99.2/96.2\\
		Leather & \textbf{100.0}/\textbf{108.3}& 99.6/99.1& 94.5/98.8& 98.2/99.2& 99.5/98.8& 99.6/98.4\\
		Tile & \textbf{96.1}/\textbf{81.3}& 90.9/77.1& 93.6/76.8& 92.5/80.8& 90.5/76.8& 90.8/73.8\\
		Wood & \textbf{98.5}/92.5& 97.8/94.0& 96.9/94.4& 98.4/\textbf{95.3}& 97.9/93.8& 98.0/93.4\\\midrule
		Bottle & \textbf{100.0}/91.1& 99.8/92.1& 98.5/92.0& 99.9/\textbf{92.6}& 99.9/88.5& 99.9/90.4\\
		Cable & \textbf{93.9}/86.5& 78.3/74.7& 90.6/84.2& 93.2/\textbf{86.6}& 81.2/67.0& 78.0/57.7\\
		Capsule & 92.6/89.4& 93.5/95.2& 93.4/95.3& \textbf{97.6}/\textbf{95.8}& 89.6/94.4& 87.7/92.1\\
		Hazelnut & 78.2/92.8& 80.1/93.9& 92.0/94.1& \textbf{97.7}/\textbf{94.2}& 88.5/92.9& 83.9/91.2\\
		Metal nut & \textbf{96.4}/84.7& 91.2/81.6& 91.6/87.9& 95.7/\textbf{89.2}& 94.6/83.8& 89.6/77.6\\
		Pill & 93.1/90.1& 93.1/\textbf{91.8}& 95.1/91.5& \textbf{97.5}/90.7& 91.9/87.4& 90.6/83.2\\
		Screw & 74.7/69.0& 65.7/84.8& 75.6/82.8& \textbf{81.2}/\textbf{88.6}& 71.4/70.2& 74.0/74.1\\
		Toothbrush & 88.1/83.5& 88.1/90.8& 97.8/90.2& \textbf{99.7}/\textbf{92.1}& 87.5/86.1& 84.2/83.7\\
		Transistor & 92.1/\textbf{87.9}& 88.5/73.2& 92.3/76.5& \textbf{96.7}/83.1& 88.0/62.9& 85.3/51.0\\
		Zipper & 94.9/92.4& 96.3/94.0& 97.3/91.9& \textbf{97.5}/\textbf{94.5}& 94.5/90.0& 94.4/90.5\\\midrule
		\textit{Average} & \makecell{ 93.1/89.1 \\ ($\pm0.29$ / $\pm 1.08$)}& \makecell{ 90.5/88.5 \\ ($\pm0.13$ / $\pm 0.35$)}& \makecell{ 91.8/89.5 \\ ($\pm0.31$ / $\pm 0.35$)}& \makecell{ \textbf{95.0}/\textbf{91.4} \\ ($\pm0.38$ / $\pm 0.27$)}& \makecell{ 91.4/85.0 \\ ($\pm0.35$ / $\pm 0.27$)}& \makecell{ 90.1/82.8 \\ ($\pm0.36$ / $\pm 0.46$)}\\

    \bottomrule
    \end{tabular}
    \caption{MVTec AD results of all 6 setups for FastFlow. The row contains results for a particular category, with columns containing detection and localization results (AUROC/AUPRO) for each setup. The mean of 3 runs is reported with the corresponding standard deviation in parentheses. The best result for each category is in \textbf{bold}.}
    \label{tab:mvtec_fastflow}
\end{table*}

\begin{table*}[!ht]
    \centering
        \setlength{\tabcolsep}{1pt}
    \begin{tabular}{lcccccc}
    \toprule
		~  & \makecell{Single model \\ 256} & \makecell{Single model \\  512} & \makecell{Tiled ensemble \\ 4 tiles} & \makecell{Tiled ensemble \\ 9 tiles} & \makecell{Single model \\ 4 tiles} & \makecell{Single model \\ 9 tiles}\\ \hline
		Carpet & 96.2/96.1& 97.2/\textbf{96.6}& 84.1/85.6& 82.5/68.6& \textbf{97.7}/96.4& 86.0/95.1\\
		Grid & 84.6/64.4& 96.1/96.0& 82.8/90.5& 90.9/93.0& \textbf{99.4}/\textbf{98.1}& 99.0/98.1\\
		Leather & \textbf{83.6}/94.7& 76.9/92.3& 49.1/73.5& 60.3/75.4& 57.7/67.6& 44.3/\textbf{95.7}\\
		Tile & 87.1/\textbf{78.7}& 75.0/64.0& \textbf{89.4}/45.9& 86.8/52.5& 88.1/67.6& 47.8/75.5\\
		Wood & \textbf{98.6}/91.0& 87.6/90.9& 97.3/\textbf{92.9}& 98.1/88.1& 89.4/88.5& 53.8/89.9\\\midrule
		Bottle & \textbf{99.8}/\textbf{95.1}& 68.8/90.9& 99.2/93.2& 98.6/92.0& 88.4/91.4& 89.5/91.2\\
		Cable & \textbf{97.8}/\textbf{91.6}& 59.4/69.8& 71.1/69.2& 82.6/77.6& 67.0/64.3& 74.1/78.2\\
		Capsule & 81.5/90.6& 71.3/93.7& 87.2/93.3& \textbf{89.7}/\textbf{93.8}& 61.0/93.2& 74.7/92.8\\
		Hazelnut & 86.1/95.2& 81.6/80.7& 94.7/73.7& \textbf{98.4}/72.6& 94.3/\textbf{96.7}& 69.6/96.2\\
		Metal nut & \textbf{100.0}/\textbf{94.3}& 90.2/88.2& 92.2/91.3& 96.8/92.8& 89.3/87.9& 82.4/87.3\\
		Pill & 90.1/94.9& 59.5/95.2& 87.7/96.9& \textbf{91.6}/\textbf{97.4}& 66.5/95.9& 89.6/96.8\\
		Screw & 75.4/92.0& 64.6/91.6& 58.8/74.9& 69.6/76.0& 72.9/93.1& \textbf{84.3}/\textbf{94.2}\\
		Toothbrush & 97.0/92.4& 97.8/\textbf{96.6}& 97.0/95.9& \textbf{99.9}/96.6& 89.0/95.5& 98.1/96.1\\
		Transistor & \textbf{95.3}/\textbf{77.1}& 73.1/62.0& 89.7/66.1& 91.2/69.0& 69.3/61.6& 80.5/60.9\\
		Zipper & \textbf{88.6}/95.0& 78.6/94.2& 88.0/\textbf{95.7}& 79.5/93.9& 79.9/95.2& 85.4/95.5\\\midrule
		\textit{Average} & \makecell{ \textbf{90.8}/\textbf{89.5} \\ ($\pm3.01$ / $\pm 3.38$)}& \makecell{ 78.5/87.2 \\ ($\pm3.25$ / $\pm 1.85$)}& \makecell{ 84.5/82.6 \\ ($\pm2.51$ / $\pm 4.99$)}& \makecell{ 87.8/82.6 \\ ($\pm1.54$ / $\pm 3.03$)}& \makecell{ 80.7/86.2 \\ ($\pm1.76$ / $\pm 0.66$)}& \makecell{ 77.3/89.5 \\ ($\pm0.92$ / $\pm 1.04$)}\\

    \bottomrule
    \end{tabular}
    \caption{MVTec AD results of all 6 setups for Reverse Distillation. The row contains results for a particular category, with columns containing detection and localization results (AUROC/AUPRO) for each setup. The mean of 3 runs is reported with the corresponding standard deviation in parentheses. The best result for each category is in \textbf{bold}.}
    \label{tab:mvtec_reverse_distillation}
\end{table*}

\begin{table*}[!ht]
    \centering
        \setlength{\tabcolsep}{1pt}
    \begin{tabular}{lcccccc}
    \toprule
		~  & \makecell{Single model \\ 256} & \makecell{Single model \\  512} & \makecell{Tiled ensemble \\ 4 tiles} & \makecell{Tiled ensemble \\ 9 tiles} & \makecell{Single model \\ 4 tiles} & \makecell{Single model \\ 9 tiles}\\ \hline
		Candle & 97.0/95.1& 98.2/96.2& 98.7/97.4& \textbf{99.3}/97.3& 98.3/97.5& 98.5/\textbf{97.6}\\
		Capsules & 71.5/70.0& \textbf{91.9}/\textbf{97.2}& 70.7/94.2& 78.1/96.0& 80.8/95.5& 90.0/96.4\\
		Cashew & 93.7/91.5& 96.5/91.5& 97.0/89.6& \textbf{97.4}/\textbf{91.6}& 94.9/90.2& 96.6/89.7\\
		Chewing gum & 98.9/84.3& 98.6/85.9& 99.1/83.8& \textbf{99.8}/83.8& 99.2/\textbf{88.5}& 98.8/84.9\\
		Fryum & 92.7/83.9& \textbf{98.4}/91.7& 93.0/90.3& 96.2/\textbf{91.9}& 94.5/90.5& 97.0/91.1\\
		Macaroni 1 & 92.3/93.5& \textbf{98.7}/96.7& 95.6/98.2& 96.0/\textbf{98.3}& 97.0/97.1& 98.7/98.1\\
		Macaroni 2 & 72.8/85.9& \textbf{91.4}/94.8& 76.4/94.0& 85.2/95.3& 87.0/94.1& 89.6/\textbf{96.2}\\
		PCB1 & 94.9/92.8& 97.9/96.5& 98.2/95.7& \textbf{98.8}/\textbf{96.5}& 98.0/95.2& 97.9/96.0\\
		PCB2 & 92.4/88.4& 97.4/93.4& 96.2/93.6& 97.5/\textbf{93.8}& \textbf{97.7}/93.6& 97.7/93.0\\
		PCB3 & \textbf{99.0}/86.6& 98.4/94.4& 94.7/93.7& 97.5/\textbf{94.6}& 97.5/93.9& 97.9/94.2\\
		PCB4 & 99.0/86.6& 99.4/\textbf{92.6}& 99.5/89.1& \textbf{99.8}/89.1& 98.5/90.8& 98.5/90.3\\
		Pipe fryum & 99.3/94.1& \textbf{99.6}/\textbf{96.9}& 98.5/96.1& 99.1/96.8& 99.6/96.1& 99.4/96.5\\\midrule
		\textit{Average} & \makecell{ 92.0/87.7 \\ ($\pm0.29$ / $\pm 0.39$)}& \makecell{ \textbf{97.2}/\textbf{94.0} \\ ($\pm0.47$ / $\pm 0.34$)}& \makecell{ 93.1/93.0 \\ ($\pm0.07$ / $\pm 0.05$)}& \makecell{ 95.4/93.7 \\ ($\pm0.81$ / $\pm 0.07$)}& \makecell{ 95.2/93.6 \\ ($\pm0.91$ / $\pm 0.55$)}& \makecell{ 96.7/93.7 \\ ($\pm0.09$ / $\pm 0.31$)}\\

    \bottomrule
    \end{tabular}
    \caption{VisA results of all 6 setups for Patchcore. The row contains results for a particular category, with columns containing detection and localization results (AUROC/AUPRO) for each setup. The mean of 3 runs is reported with the corresponding standard deviation in parentheses. The best result for each category is in \textbf{bold}.}
    \label{tab:visa_patchcore}
\end{table*}

\begin{table*}[!ht]
    \centering
        \setlength{\tabcolsep}{1pt}
    \begin{tabular}{lcccccc}
    \toprule
		~  & \makecell{Single model \\ 256} & \makecell{Single model \\  512} & \makecell{Tiled ensemble \\ 4 tiles} & \makecell{Tiled ensemble \\ 9 tiles} & \makecell{Single model \\ 4 tiles} & \makecell{Single model \\ 9 tiles}\\ \hline
		Candle & 87.4/92.3& 86.0/\textbf{96.1}& 90.1/96.1& \textbf{94.0}/92.3& 85.8/96.1& 86.5/96.1\\
		Capsules & 60.5/58.7& 62.3/\textbf{76.4}& 67.9/74.2& \textbf{67.9}/75.5& 61.9/74.2& 62.3/75.4\\
		Cashew & \textbf{88.8}/\textbf{84.7}& 87.3/83.1& 88.6/83.6& 88.4/83.6& 87.6/83.7& 87.6/83.2\\
		Chewing gum & \textbf{98.4}/83.7& 97.5/84.2& 96.6/83.8& 97.2/84.3& 89.2/\textbf{87.5}& 97.4/83.8\\
		Fryum & 86.4/77.4& 88.8/86.6& 87.9/86.9& \textbf{89.6}/86.7& 88.2/\textbf{87.0}& 88.5/86.6\\
		Macaroni 1 & \textbf{80.3}/89.2& 76.7/\textbf{91.4}& 75.6/91.3& 79.7/91.3& 76.6/91.3& 77.2/91.3\\
		Macaroni 2 & 71.8/76.4& 66.3/72.3& 70.3/77.3& \textbf{72.2}/\textbf{78.3}& 66.2/77.3& 66.2/77.8\\
		PCB1 & 88.6/88.5& 84.7/91.9& 87.5/91.9& \textbf{89.9}/92.1& 84.9/91.9& 80.3/\textbf{93.2}\\
		PCB2 & 81.7/83.9& 79.3/90.7& 79.5/90.6& \textbf{84.4}/\textbf{90.9}& 80.4/90.7& 80.0/90.8\\
		PCB3 & 72.5/80.5& 73.8/89.6& 75.5/89.5& \textbf{80.7}/89.7& 74.3/89.4& 74.1/\textbf{89.8}\\
		PCB4 & 96.4/81.7& 87.8/\textbf{88.5}& 96.5/85.3& \textbf{97.1}/85.0& 95.8/85.3& 95.3/84.7\\
		Pipe fryum & 92.3/88.3& 92.1/92.4& 89.8/92.2& \textbf{94.6}/\textbf{92.5}& 91.2/92.2& 91.7/92.2\\\midrule
		\textit{Average} & \makecell{ 83.7/82.1 \\ ($\pm0.90$ / $\pm 0.87$)}& \makecell{ 81.9/86.9 \\ ($\pm1.76$ / $\pm 0.81$)}& \makecell{ 83.8/86.9 \\ ($\pm0.96$ / $\pm 0.84$)}& \makecell{ \textbf{86.3}/86.9 \\ ($\pm0.96$ / $\pm 1.29$)}& \makecell{ 81.8/\textbf{87.2} \\ ($\pm1.96$ / $\pm 1.12$)}& \makecell{ 82.3/87.1 \\ ($\pm0.58$ / $\pm 0.89$)}\\

    \bottomrule
    \end{tabular}
    \caption{VisA results of all 6 setups for Padim. The row contains results for a particular category, with columns containing detection and localization results (AUROC/AUPRO) for each setup. The mean of 3 runs is reported with the corresponding standard deviation in parentheses. The best result for each category is in \textbf{bold}.}
    \label{tab:visa_padim}
\end{table*}

\begin{table*}[!ht]
    \centering
        \setlength{\tabcolsep}{1pt}
    \begin{tabular}{lcccccc}
    \toprule
		~  & \makecell{Single model \\ 256} & \makecell{Single model \\  512} & \makecell{Tiled ensemble \\ 4 tiles} & \makecell{Tiled ensemble \\ 9 tiles} & \makecell{Single model \\ 4 tiles} & \makecell{Single model \\ 9 tiles}\\ \hline
		Candle & 90.7/92.4& 92.8/96.5& 92.4/96.3& \textbf{93.8}/\textbf{96.6}& 92.5/96.1& 90.9/94.8\\
		Capsules & 77.5/81.2& 84.1/\textbf{92.9}& 80.4/87.3& \textbf{85.3}/92.4& 78.7/86.0& 80.8/83.7\\
		Cashew & 88.9/82.6& 88.5/\textbf{90.2}& 91.3/88.0& \textbf{92.8}/90.0& 87.3/83.5& 87.6/79.6\\
		Chewing gum & 98.7/85.6& 99.6/\textbf{90.4}& \textbf{99.8}/88.7& 99.7/89.2& 99.0/88.2& 98.4/85.5\\
		Fryum & 93.8/74.7& 94.2/\textbf{81.2}& 93.7/76.6& \textbf{97.2}/79.6& 91.6/75.2& 89.1/65.0\\
		Macaroni 1 & 87.7/88.1& 90.6/91.8& 85.3/91.5& \textbf{91.0}/\textbf{94.2}& 88.6/91.4& 89.0/87.6\\
		Macaroni 2 & \textbf{73.2}/82.5& 72.5/\textbf{88.0}& 68.3/83.3& 70.4/83.7& 69.1/82.2& 67.3/80.2\\
		PCB1 & 85.5/86.4& 88.5/90.0& 91.1/88.5& \textbf{95.0}/\textbf{93.5}& 83.7/83.8& 81.9/74.3\\
		PCB2 & 83.3/75.4& 88.5/83.6& 87.9/84.2& \textbf{94.2}/\textbf{88.3}& 81.9/75.7& 82.5/70.6\\
		PCB3 & 79.1/65.2& 86.8/85.2& 89.2/86.0& \textbf{93.8}/\textbf{88.1}& 83.8/78.2& 80.0/72.2\\
		PCB4 & 95.3/80.0& 97.5/82.4& 99.0/83.9& \textbf{99.4}/\textbf{83.9}& 94.4/78.0& 86.3/69.7\\
		Pipe fryum & 95.3/83.0& 94.3/88.0& 94.5/87.6& \textbf{97.9}/\textbf{90.2}& 94.9/82.3& 93.3/80.2\\\midrule
		\textit{Average} & \makecell{ 87.4/81.4 \\ ($\pm0.42$ / $\pm 1.10$)}& \makecell{ 89.8/88.4 \\ ($\pm0.27$ / $\pm 0.22$)}& \makecell{ 89.4/86.8 \\ ($\pm0.31$ / $\pm 0.34$)}& \makecell{ \textbf{92.5}/\textbf{89.2} \\ ($\pm0.12$ / $\pm 0.13$)}& \makecell{ 87.1/83.4 \\ ($\pm0.32$ / $\pm 0.70$)}& \makecell{ 85.6/78.6 \\ ($\pm0.18$ / $\pm 2.62$)}\\

    \bottomrule
    \end{tabular}
    \caption{VisA results of all 6 setups for FastFlow. The row contains results for a particular category, with columns containing detection and localization results (AUROC/AUPRO) for each setup. The mean of 3 runs is reported with the corresponding standard deviation in parentheses. The best result for each category is in \textbf{bold}.}
    \label{tab:visa_fastflow}
\end{table*}

\begin{table*}[!ht]
    \centering
        \setlength{\tabcolsep}{1pt}
    \begin{tabular}{lcccccc}
    \toprule
		~  & \makecell{Single model \\ 256} & \makecell{Single model \\  512} & \makecell{Tiled ensemble \\ 4 tiles} & \makecell{Tiled ensemble \\ 9 tiles} & \makecell{Single model \\ 4 tiles} & \makecell{Single model \\ 9 tiles}\\ \hline
		Candle & 88.2/93.7& 60.4/96.7& 89.7/\textbf{97.2}& \textbf{93.0}/96.9& 78.4/97.0& 84.9/96.5\\
		Capsules & 76.9/88.3& 74.2/87.4& 80.2/94.8& 84.0/\textbf{96.0}& 75.2/93.3& \textbf{87.6}/95.5\\
		Cashew & 92.4/80.9& 88.7/\textbf{83.6}& 93.7/75.8& \textbf{94.4}/75.2& 72.2/76.8& 89.0/77.8\\
		Chewing gum & 98.4/\textbf{85.0}& 69.5/68.1& \textbf{98.9}/69.9& 96.3/64.3& 78.2/71.2& 95.8/70.8\\
		Fryum & 81.0/82.7& 53.7/89.4& 82.8/88.1& \textbf{92.4}/\textbf{90.0}& 52.1/88.9& 82.3/86.8\\
		Macaroni 1 & 78.4/87.4& 88.0/\textbf{96.3}& 85.8/96.2& 88.0/96.1& \textbf{88.6}/96.3& 81.5/95.2\\
		Macaroni 2 & 59.5/82.1& \textbf{72.9}/\textbf{93.7}& 72.0/92.9& 72.7/93.3& 72.3/92.2& 69.5/92.1\\
		PCB1 & 64.7/94.0& 59.3/93.5& 91.1/94.1& \textbf{93.3}/\textbf{95.4}& 62.8/93.1& 62.2/92.4\\
		PCB2 & 89.0/87.1& 45.3/89.3& 86.6/91.0& \textbf{92.0}/\textbf{92.7}& 60.8/88.2& 69.4/89.1\\
		PCB3 & 81.3/85.9& 85.2/90.9& 85.8/91.2& \textbf{94.6}/\textbf{93.3}& 75.0/89.2& 82.6/90.4\\
		PCB4 & 95.9/83.7& 84.6/87.6& 97.7/88.5& \textbf{99.3}/\textbf{91.7}& 64.6/86.7& 73.9/85.9\\
		Pipe fryum & 93.1/92.2& 76.3/\textbf{95.6}& 95.7/95.2& \textbf{96.7}/95.0& 90.4/94.7& 88.3/94.7\\\midrule
		\textit{Average} & \makecell{ 83.2/86.9 \\ ($\pm1.93$ / $\pm 0.66$)}& \makecell{ 71.5/89.3 \\ ($\pm5.96$ / $\pm 1.70$)}& \makecell{ 88.3/89.6 \\ ($\pm1.54$ / $\pm 0.12$)}& \makecell{ \textbf{91.4}/\textbf{90.0} \\ ($\pm1.32$ / $\pm 0.47$)}& \makecell{ 72.5/89.3 \\ ($\pm4.29$ / $\pm 0.48$)}& \makecell{ 80.6/88.9 \\ ($\pm2.59$ / $\pm 1.03$)}\\

    \bottomrule
    \end{tabular}
    \caption{VisA results of all 6 setups for Reverse Distillation. The row contains results for a particular category, with columns containing detection and localization results (AUROC/AUPRO) for each setup. The mean of 3 runs is reported with the corresponding standard deviation in parentheses. The best result for each category is in \textbf{bold}.}
    \label{tab:visa_reverse_distillation}
\end{table*}

\FloatBarrier

\section{Additional qualitative examples.}
\label{ap:qual_res}

This section contains qualitative examples for every architecture and every setup on all categories of both MVTec AD and VisA.
\begin{figure*}[!b]
    \centering
    \includegraphics[width=1\linewidth]{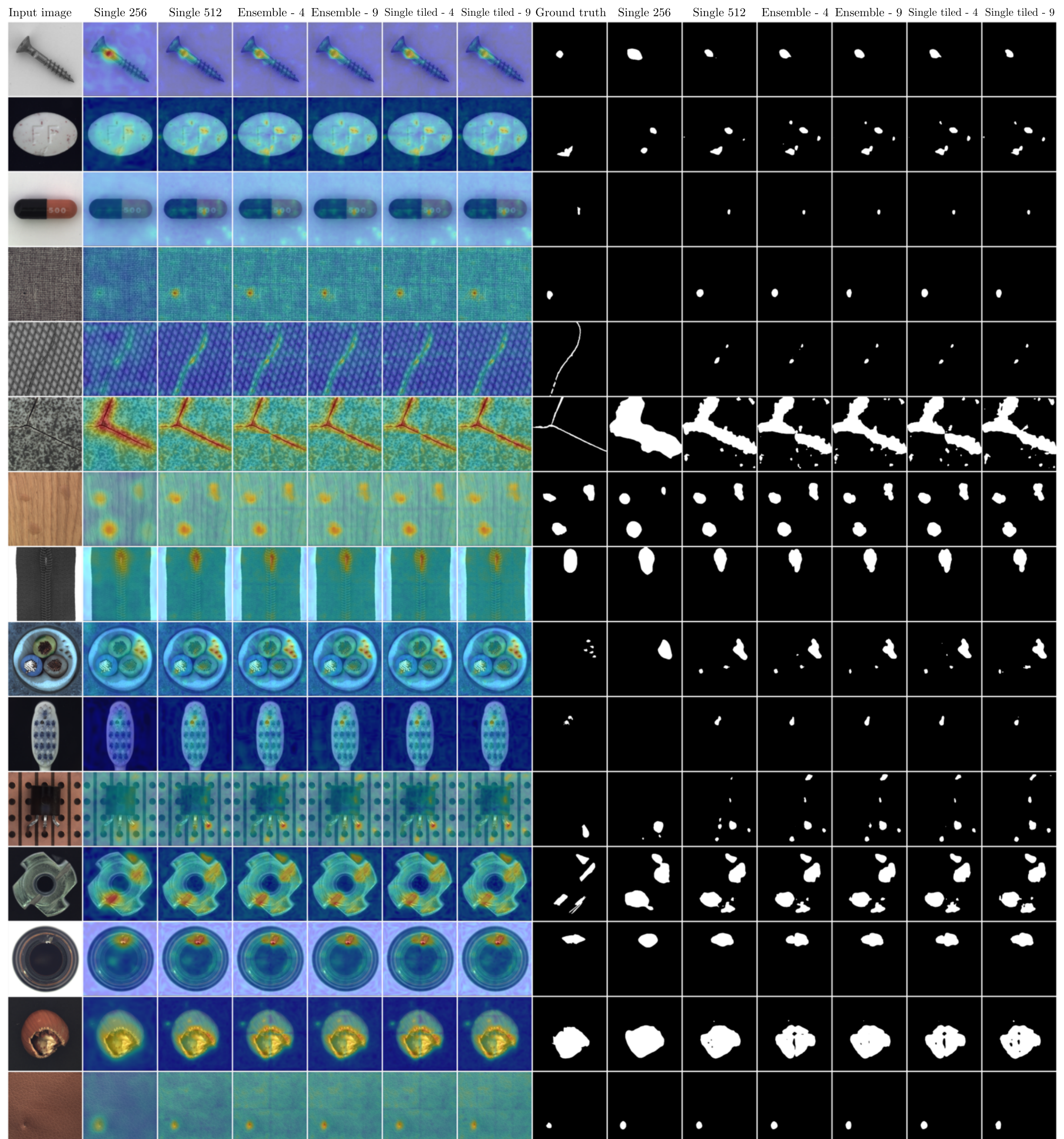}
    \caption{Anomaly maps and segmentation masks for each setup using PatchCore on randomly picked defective image from every category in MVTec AD.}
    \label{fig:qualit_pc_mvtec}
\end{figure*}

\begin{figure*}[!t]
    \centering
    \includegraphics[width=1\linewidth]{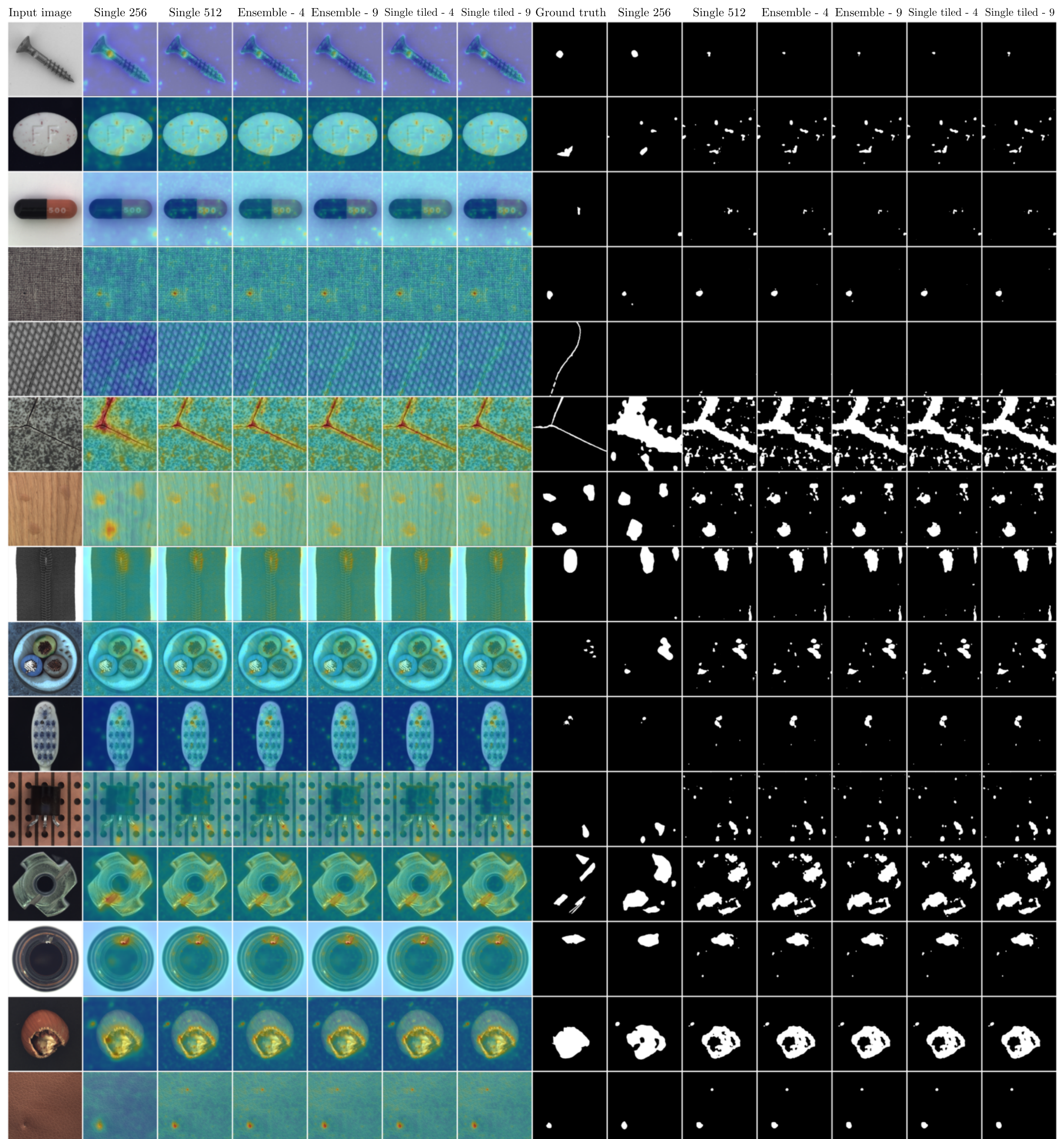}
    \caption{Anomaly maps and segmentation masks for each setup using Padim on randomly picked defective image from every category in MVTec AD.}
    \label{fig:qualit_pd_mvtec}
\end{figure*}

\begin{figure*}[!t]
    \centering
    \includegraphics[width=1\linewidth]{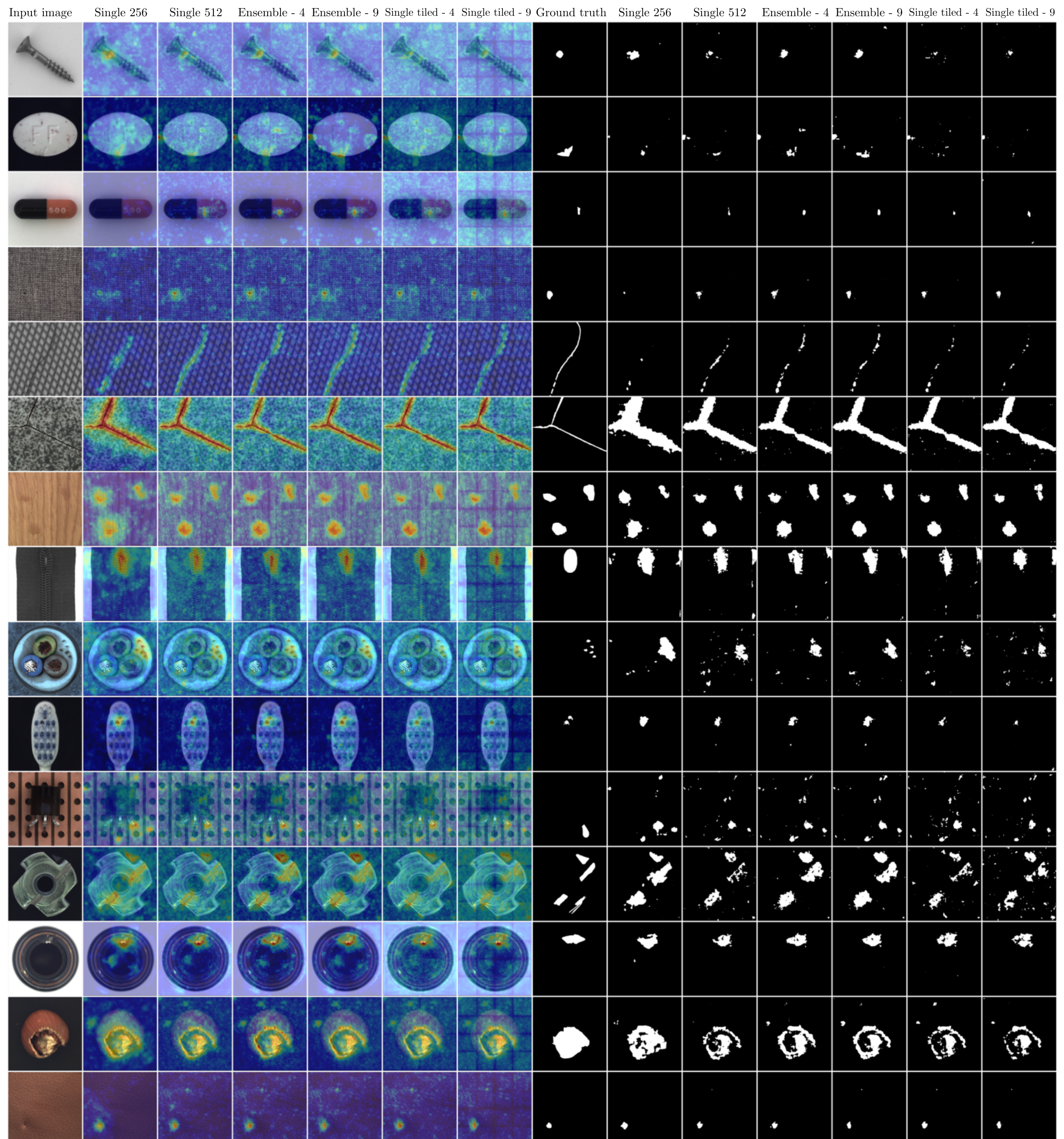}
    \caption{Anomaly maps and segmentation masks for each setup using FastFlow on randomly picked defective image from every category in MVTec AD.}
    \label{fig:qualit_ff_mvtec}
\end{figure*}

\begin{figure*}[!t]
    \centering
    \includegraphics[width=1\linewidth]{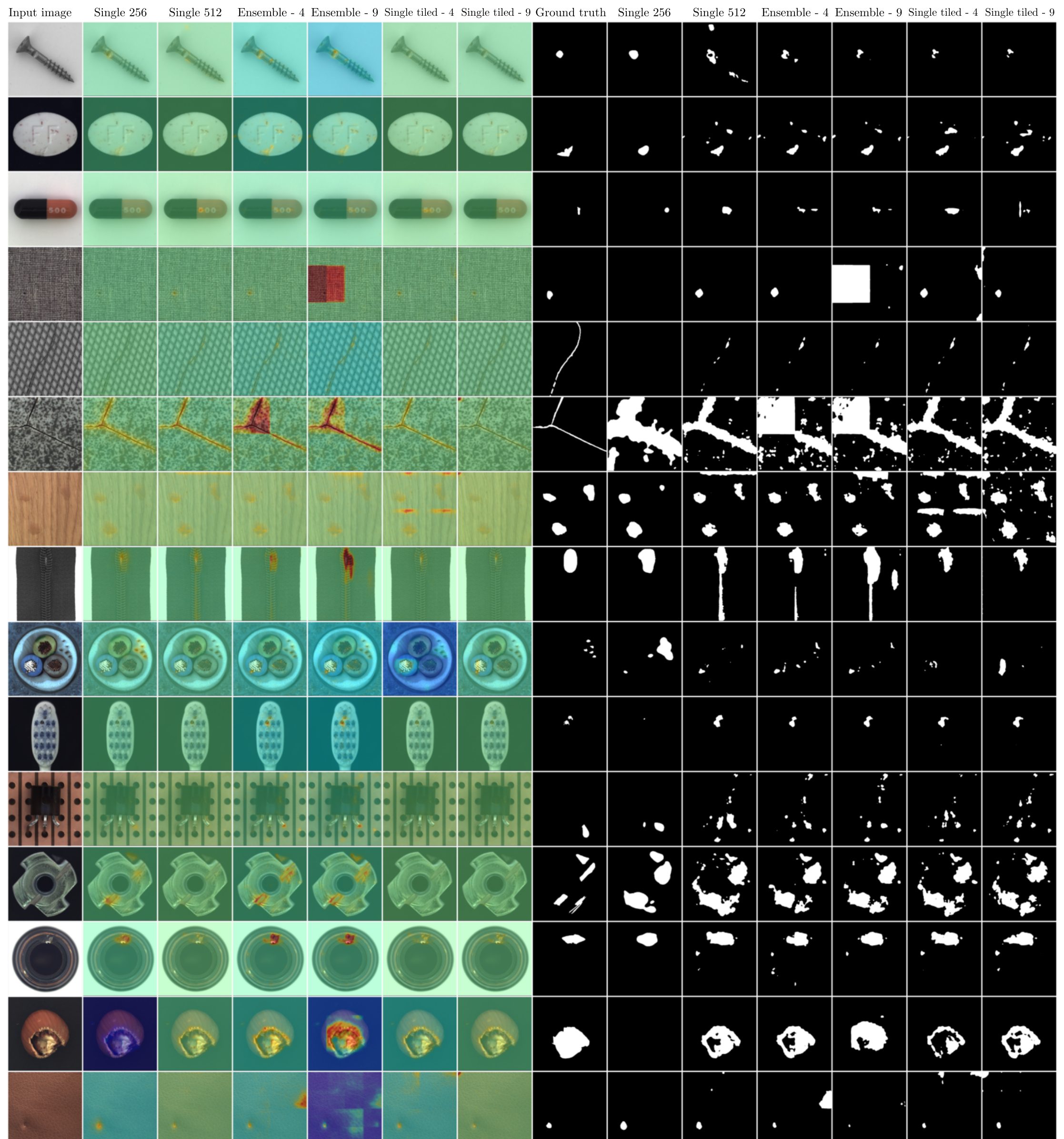}
    \caption{Anomaly maps and segmentation masks for each setup using Reverse Distillation on randomly picked defective image from every category in MVTec AD.}
    \label{fig:qualit_rd_mvtec}
\end{figure*}

\begin{figure*}[!t]
    \centering
    \includegraphics[width=1\linewidth]{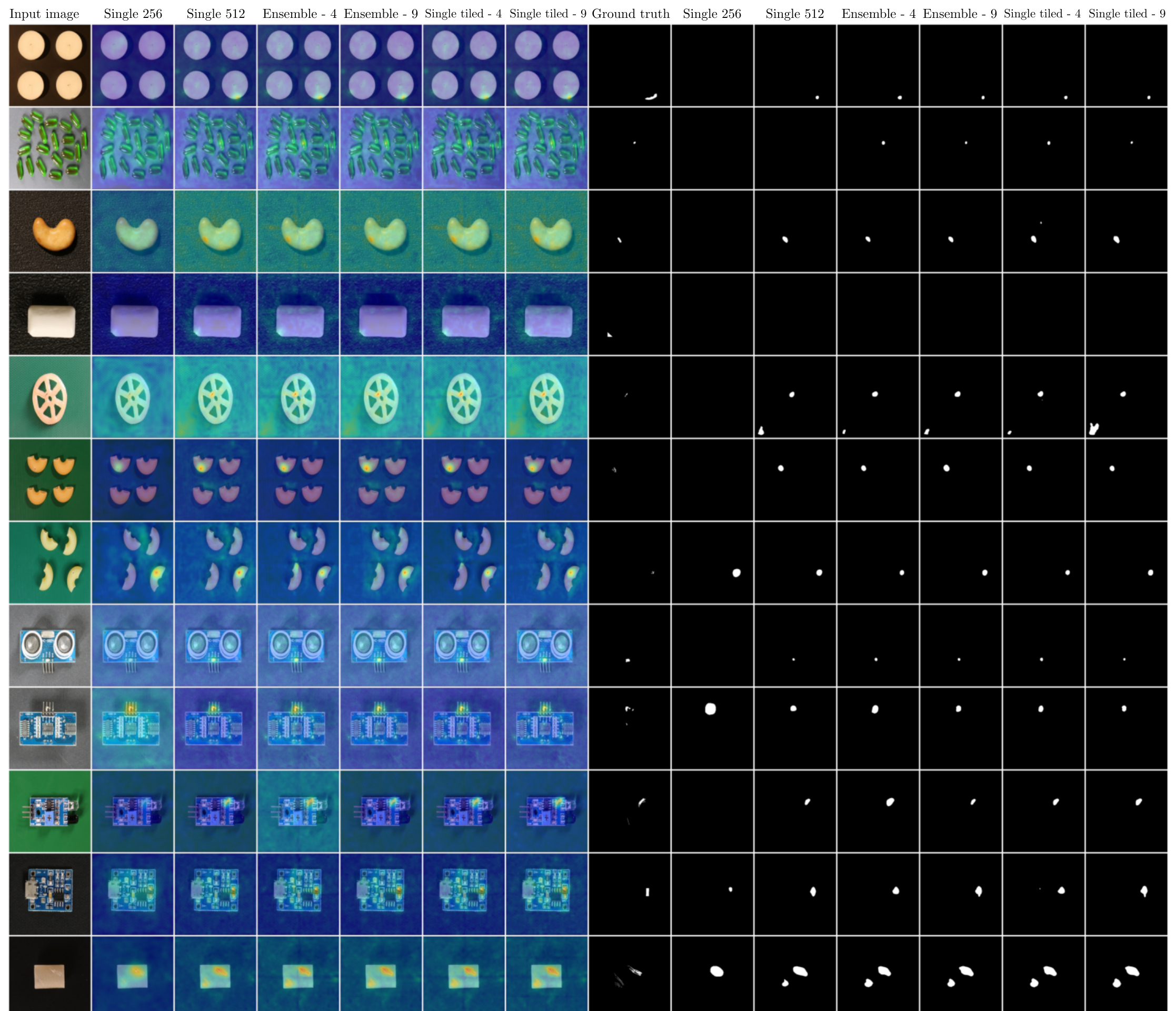}
    \caption{Anomaly maps and segmentation masks for each setup using PatchCore on randomly picked defective image from every category in VisA.}
    \label{fig:qualit_pc_visa}
\end{figure*}

\begin{figure*}[!t]
    \centering
    \includegraphics[width=1\linewidth]{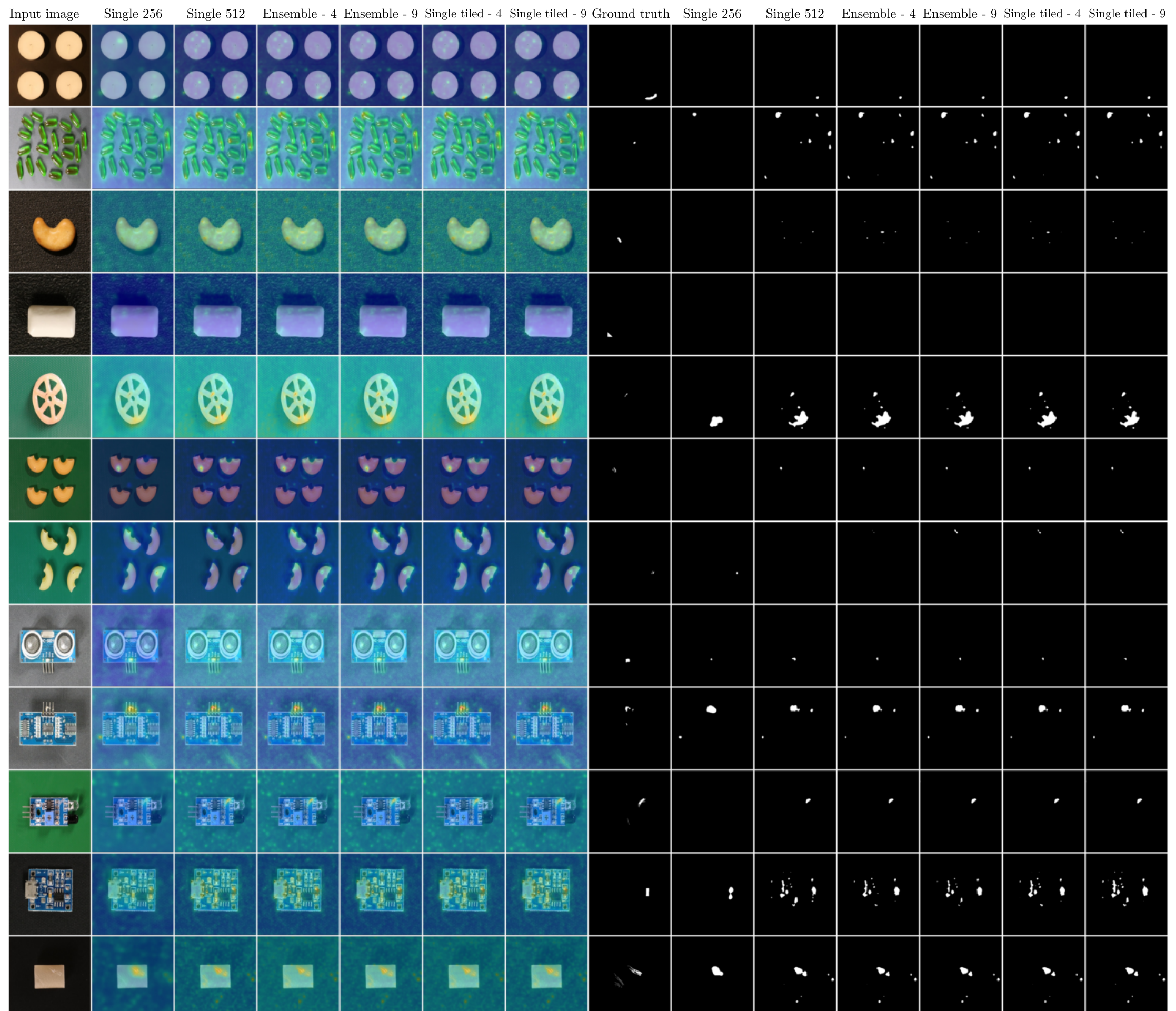}
    \caption{Anomaly maps and segmentation masks for each setup using Padim on randomly picked defective image from every category in VisA.}
    \label{fig:qualit_pd_visa}
\end{figure*}

\begin{figure*}[!t]
    \centering
    \includegraphics[width=1\linewidth]{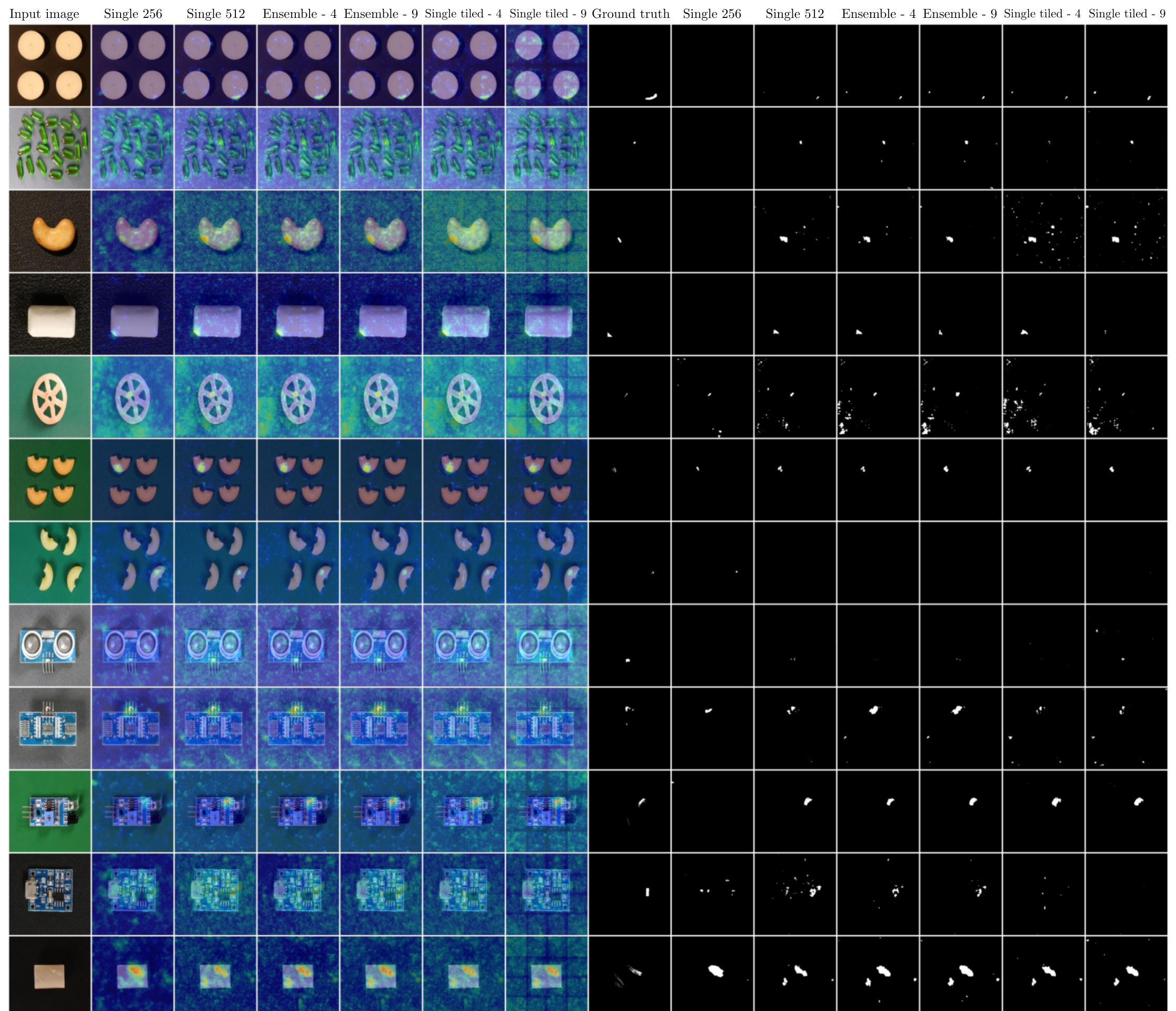}
    \caption{Anomaly maps and segmentation masks for each setup using FastFlow on randomly picked defective image from every category in VisA.}
    \label{fig:qualit_ff_visa}
\end{figure*}

\begin{figure*}[!t]
    \centering
    \includegraphics[width=1\linewidth]{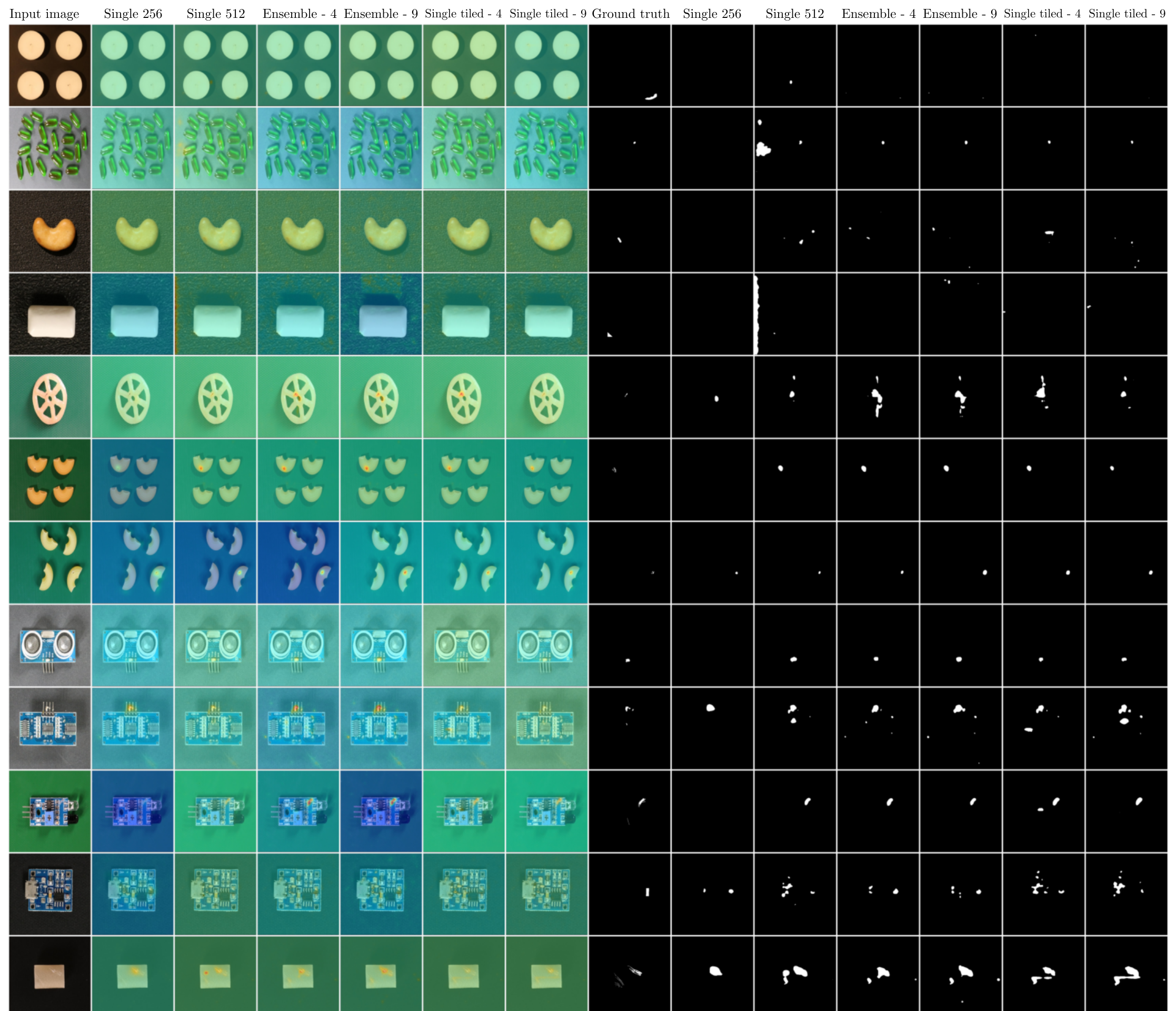}
    \caption{Anomaly maps and segmentation masks for each setup using Reverse Distillation on randomly picked defective image from every category in VisA.}
    \label{fig:qualit_red_visa}
\end{figure*}